\UseRawInputEncoding
\documentclass[runningheads]{llncs}

\usepackage{eccv}

\usepackage{eccvabbrv}

\usepackage{graphicx}
\usepackage{booktabs}

\usepackage[accsupp]{axessibility}  

\usepackage[pagebackref,breaklinks,colorlinks,citecolor=eccvblue]{hyperref}
\usepackage{paralist}
\usepackage{array}
\usepackage{multirow}
\usepackage{amsmath}
\usepackage{colortbl}

\newcommand{\mypartitle}[2][2.25]{\vspace*{-#1 ex}~\\{\noindent {\bf #2}}}

\newcolumntype{P}[1]{>{\centering\arraybackslash}p{#1}}
\newcolumntype{M}[1]{>{\centering\arraybackslash}m{#1}}
\newcolumntype{L}[1]{>{\arraybackslash}m{#1}}

\newcommand{\LAH}{LAH\xspace}
\newcommand{\LAEO}{LAEO\xspace}
\newcommand{\SA}{SA\xspace}

\newcommand{\inputimage}{\ensuremath{\mathbf{I}}\xspace}
\newcommand{\inputheadbbox}{\ensuremath{\mathbf{h}^{\text{box}}}\xspace}
\newcommand{\HeadCrop}{\ensuremath{\mathbf{h}^{\text{crop}}}\xspace}
\newcommand{\HeadSeqCrop}{\ensuremath{\mathbf{H}^{\text{img}}}\xspace}
\newcommand{\HeadSeqBbox}{\ensuremath{\mathbf{H}^{\text{box}}}\xspace}

\newcommand{\Npersons}{\ensuremath{N_p}\xspace}

\newcommand{\Ntimesteps}{\ensuremath{T}\xspace}

\newcommand{\imgtokens}{\ensuremath{\mathbf{f}\xspace}}
\newcommand{\persontoken}{\ensuremath{\mathbf{p}\xspace}}
\newcommand{\gazetoken}{\ensuremath{\mathbf{g}\xspace}}
\newcommand{\gazetokenstatic}{\ensuremath{\gazetoken^{\text{stat}}}\xspace}
\newcommand{\gazetokentemp}{\ensuremath{\gazetoken^{\text{temp}}}\xspace}
\newcommand{\scenepersontoken}{\ensuremath{\persontoken^{\text{s,b}}}\xspace}
\newcommand{\socialpersontoken}{\ensuremath{\persontoken^{\text{p,b}}}\xspace}
\newcommand{\inpersontoken}{\ensuremath{\persontoken^{\text{o,b-1}}}\xspace}
\newcommand{\outpersontoken}{\ensuremath{\persontoken^{\text{o,b}}}\xspace}
\newcommand{\mspersontoken}{\ensuremath{\persontoken^{\text{ms}}}\xspace}
\newcommand{\inimgtokens}{\ensuremath{\imgtokens^{\text{o,b-1}}}\xspace}
\newcommand{\peopleimgtokens}{\ensuremath{\imgtokens^{\text{p,b}}}\xspace}
\newcommand{\outimgtokens}{\ensuremath{\imgtokens^{\text{o,b}}}\xspace}

\newcommand{\gazebackbone}{\ensuremath{\mathcal{G}_\text{stat}}\xspace}
\newcommand{\gazetempencoder}{\ensuremath{\mathcal{G}_\text{temp}}\xspace}
\newcommand{\gazevectorMLP}{\ensuremath{\mathcal{G}_\text{vec}}\xspace}

\newcommand{\persontoscene}{\ensuremath{\mathcal{I}_\text{ps}^\text{b}}\xspace}
\newcommand{\vit}{\ensuremath{\mathcal{V}}\xspace}
\newcommand{\vitlayer}{\ensuremath{l}\xspace}
\newcommand{\scenetoperson}{\ensuremath{\mathcal{I}_\text{sp}^\text{b}}\xspace}
\newcommand{\socialencoder}{\ensuremath{\mathcal{I}_\text{pp}^\text{b}}\xspace}
\newcommand{\persontempencoder}{\ensuremath{\mathcal{I}_\text{pt}^\text{b}}\xspace}

\newcommand{\lahMLP}{\ensuremath{E}\xspace}    
\newcommand{\coattMLP}{\ensuremath{C}\xspace}    
\newcommand{\inoutMLP}{\ensuremath{\mathcal{O}}\xspace}

\newcommand{\gazevector}{\ensuremath{\mathbf{g}^{\text{v}}}\xspace}
\newcommand{\speakingstatus}{\ensuremath{\mathbf{s}\xspace}}
\newcommand{\inoutlabel}{\ensuremath{\mathbf{o}}\xspace}

\newcommand{\lookinglabel}{\ensuremath{\mathbf{e}}\xspace}    
\newcommand{\coattlabel}{\ensuremath{\mathbf{c}}\xspace}    

\newcommand{\bboxproj}{\ensuremath{\mathcal{P}_{\text{bbox}}}\xspace}
\newcommand{\gazeproj}{\ensuremath{\mathcal{P}_{\text{gaze}}}\xspace}
\newcommand{\speakingproj}{\ensuremath{\mathcal{P}_{\text{spk}}}\xspace}

\newcommand{\personproj}{\ensuremath{\mathcal{P}_{\text{social}}}\xspace}
\newcommand{\dptproj}{\ensuremath{\mathcal{P}_{\text{DPT}}}\xspace}

\newcommand{\loss}{\ensuremath{\mathcal{L}}\xspace}
\newcommand{\losscoeff}{\ensuremath{\mathcal{\lambda}}\xspace}
\newcommand{\lossheatmap}{\loss_{\text{HM}}\xspace}
\newcommand{\losscoeffheatmap}{\losscoeff_{\text{HM}}\xspace}
\newcommand{\lossangular}{\loss_{\text{VEC}}\xspace}
\newcommand{\losscoeffangular}{\losscoeff_{\text{VEC}}\xspace}
\newcommand{\losslah}{\loss_{\text{\LAH}}\xspace}
\newcommand{\losscoefflah}{\losscoeff_{\text{\LAH}}\xspace}

\newcommand{\losscoatt}{\loss_{\text{\SA}}\xspace}
\newcommand{\losscoeffcoatt}{\losscoeff_{\text{\SA}}\xspace}
\newcommand{\lossio}{\loss_{\text{IO}}\xspace}
\newcommand{\losscoeffio}{\losscoeff_{\text{IO}}\xspace}

\usepackage{orcidlink}

\usepackage{pifont}

\newcommand{\cmark}{\ding{51}}%
\newcommand{\xmark}{\ding{55}}%
\definecolor{BackgroundColor}{RGB}{247, 217, 250}

\begin{document}

\title{A Novel Framework for Multi-Person Temporal Gaze Following and Social Gaze Prediction} 

\titlerunning{MTGS}

\author{Anshul Gupta \and
Samy Tafasca \and
Arya Farkhondeh \and
Pierre Vuillecard \and
Jean-Marc Odobez}

\authorrunning{Gupta et al.}

\institute{Idiap Research Institite, Martigny, Switzerland \and
Ecole Polytechnique Federale de Lausanne, Switzerland}

\maketitle

\begin{figure}
\vspace{-4mm}
    \centering
    \captionsetup{type=figure}
    \includegraphics[width=1.\textwidth]{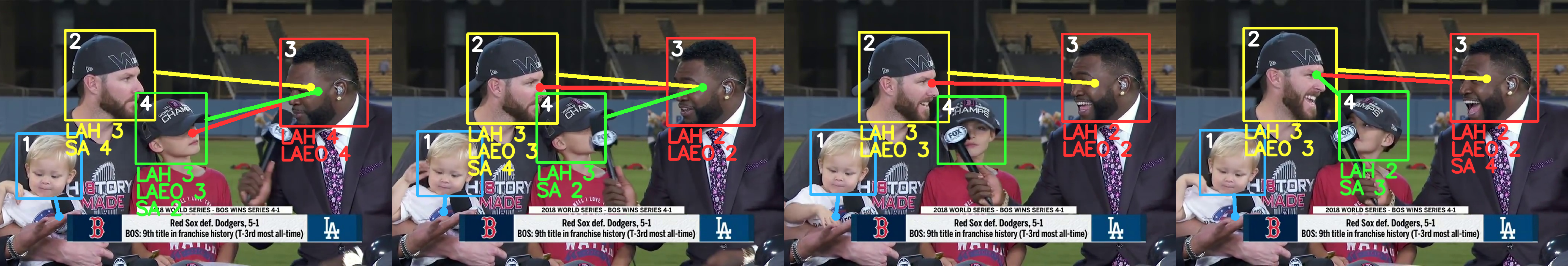}
    \captionof{figure}{Results of our proposed multi-person 
    and temporal transformer architecture for joint gaze following and social gaze prediction, namely Looking at Humans (\LAH), Looking at Each Other (\LAEO), and Shared Attention (\SA). For each person, the social gaze task is listed with the associated person ID (\emph{e.g.} in frame 1 for person 2, they are in \SA with person 4). More qualitative results can be found in the supplementary. 
}
\vspace{-6mm}
\label{fig:qualitative}
\end{figure}

\begin{abstract}
Gaze following and social gaze prediction are fundamental tasks providing insights into human communication behaviors, intent, and social interactions. Most previous approaches addressed these tasks separately, either by designing highly specialized social gaze models that do not generalize to other social gaze tasks or by considering social gaze inference as an ad-hoc post-processing of the gaze following task. Furthermore, the vast majority of gaze following approaches have proposed static models that can handle only one person at a time, therefore failing to take advantage of social interactions and temporal dynamics. In this paper, we address these limitations and introduce a novel framework to jointly predict the gaze target and social gaze label for all people in the scene. The framework comprises of: (i) a temporal, transformer-based architecture that, in addition to image tokens, handles person-specific tokens capturing the gaze information related to each individual; (ii) a new dataset, VSGaze, that unifies annotation types across multiple gaze following and social gaze datasets. We show that our model trained on VSGaze can address all tasks jointly, and achieves state-of-the-art results for multi-person gaze following and social gaze prediction. 
 \vspace{-3mm}
\end{abstract}

\vspace{-1mm}

\section{Introduction}
\label{sec:intro}

Social interaction plays a pivotal role in our daily lives and is influenced by an array of behavioral elements, encompassing not only verbal communication but also non-verbal cues such as gestures or body language.
In particular, the ability to decode people's gaze, including communicative behaviors like eye contact and shared attention on a particular object, is highly related to our capacity to connect with and pay attention to others. 
%
As such, designing social gaze prediction algorithms and systems has attracted considerable attention from different communities, ranging from medical diagnosis to human-robot interactions~\cite{sheikhi2015combining, Otsuka2018}.

%

In this work, we investigate whether we can build a unified framework to infer from video data the \textit{gaze target} and \textit{social gaze label} in \textit{one stage} for \textit{all people} in the scene. This requires: (i) A new architecture capable of jointly modeling these tasks, (ii) A large-scale dataset with annotations for all the tasks. Specifically, as social gaze tasks, we focus on \LAH, \LAEO, and \SA as introduced in Figure~\ref{fig:qualitative}. 


%
Methods for social gaze prediction in the literature adopt one of two approaches.
The first one focuses on the design of dedicated networks  
to process pairs of head crops and potentially other scene information \cite{Marin-Jimenez_2019_CVPR, marin21pami,Doosti_Chen_Vemulapalli_Jia_Zhu_Green_2021, cantarini2021hhpnet,fan2018inferring_videocoatt,sumer2020attention}. 
While their specialization makes them effective, 
they offer little room for generalization to other gaze-related tasks.
The second direction first predicts people's gaze targets, also called gaze following~\cite{nips15_recasens},
and then leverages the predicted gaze points to infer social gaze through an ad-hoc post-processing scheme ~\cite{guo2022mgtr,chong2020dvisualtargetattention, tu2022end}.
%
%

However, gaze following itself is a challenging task.
Besides geometric aspects, the task requires understanding and establishing a correspondence between top-down information related to the person's state, activity, and cognitive intent, and bottom-up saliency related to the scene context (where are salient items, are people interacting, \emph{etc.}).
%
Furthermore, gaze following performance, as measured by distance, does not always translate to similar semantic performance (\emph{e.g.} when evaluating if the predicted gaze point falls on a person's head or not~\cite{tafasca2023childplay}). 

\mypartitle{Motivation.}
%
Existing methods for gaze following suffer from several drawbacks. 
First, most of the methods perform prediction for a single person \cite{nips15_recasens,recasens2017following,chong2020dvisualtargetattention,Fang_2021_CVPR_DAM,gupta2022modular,tafasca2023childplay}, requiring multiple inference passes on the same image to process multiple people in the scene.
%
%
In contrast, a multi-person gaze following architecture processes the image only once and has to capture salient items for all people in the scene,
while maintaining the ability to infer the gaze target of each individual. This is inherently more complex and challenging.
%
Another drawback of the single-person 
formulation is that it does not explicitly model people's interactions, thereby 
preventing the possibility of jointly inferring people's gaze target and social gaze attributes. Secondly, the majority of proposed models for gaze following are static, using only a single image at a time. This is partly due to the absence of large and diverse video datasets, and the difficulty of leveraging large-scale static ones like GazeFollow~\cite{recasens2017following}. 
This is a limitation, as temporal information can capture head and gaze coordination patterns \cite{sheikhi2015combining} 
which can help gaze direction inference, especially when the eyes are not completely visible  \cite{Nonaka_2022_CVPR}.

Finally, none of these methods have investigated learning the gaze following and social gaze prediction tasks jointly. Thus, it remains a research question whether such formulation could improve performance by having social cues inform the gaze following task, or if performance would degrade as we try to accommodate multiple tasks, datasets, and people within the same framework.

\mypartitle{Contributions.}
Given these motivations, we propose a new, unified framework for gaze following and social gaze prediction with the following contributions:
\begin{compactitem}
    \item A novel temporal and multi-person architecture for gaze following and social gaze prediction. Our approach posits people as specific tokens that can interact with each other and the scene content (\emph{i.e.} image tokens). This token-based multi-person representation allows for the modeling of (i) temporal information at multiple levels (from 2D gaze direction to 2D gaze target level), (ii) the joint prediction of the gaze target and social gaze label. 
    \item A new dataset, VSGaze, that 
    unifies annotation types across multiple gaze following and social gaze datasets.
    \item New social gaze protocols and metrics for better evaluating semantic gaze following performance.
\end{compactitem}
Our architecture achieves state-of-the-art results for multi-person gaze following, while also performing competitively against single-person models. It is also able to leverage our proposed VSGaze dataset to jointly tackle gaze following and social gaze prediction, 
achieving competitive performance compared to methods trained on individual tasks. Our experiments further demonstrate that the performance benefits from this joint prediction, \emph{i.e.} adding the social loss, improves gaze following performance, and vice-versa. Finally, the new social gaze metrics provide complementary information to the standard distance-based metrics. This helps assess model performance from the perspective of social interaction.

It is worth noting that our architecture design is easily extendable and allows for the integration of auxiliary person-specific information that can influence the final predictions. In the supplementary, we explore this aspect by integrating people's speaking status in the person tokens to improve the results.

\section{Related work}
\label{sec:RelatedWork}


\mypartitle{Gaze Following.} 
Typical methods for this task 
exploit a two-branch architecture: one for processing the scene and the other for processing the person of interest \cite{nips15_recasens, chong2020dvisualtargetattention, Fang_2021_CVPR_DAM, gupta2022modular, jin2021multi, jin2022depth, lian2018believe}. They have distinguished themselves 
by the addition of other relevant modalities like depth \cite{Fang_2021_CVPR_DAM, gupta2022modular, jin2022depth}, pose \cite{gupta2022modular}, and objects \cite{hu2022gaze}, or by potentially leveraging scene geometry \cite{hu2022we, jin2022depth, Fang_2021_CVPR_DAM}.
However, only a few efforts have addressed the multi-person case.
\cite{jin2021multi} first proposed a 
simple architecture relying on a scene backbone to get a person-agnostic image representation that is
subsequently fused with the head crop features of each individual obtained using another backbone.
While this reduces computation, the model does not account for 
person interactions, as each head is processed separately. 
In another direction, \cite{tu2022end,Tonini_2023_ICCV} 
rely on a transformer-based architecture to perform multi-person gaze following. 
Their methods borrow from DETR \cite{carion2020end}, 
taking the image as input and simultaneously predicting 
the head bounding box and gaze target for every person in the scene.
While these methods can implicitly model person interactions, 
an important limitation is that they compute performance on detected heads which are matched to the ground truth. Given that both the head detection and matching steps are error-prone, it precludes comparing their results to others.  
In addition, with their formulation, adding person-specific auxiliary 
information or modeling temporal information is challenging. 

\mypartitle{Temporal gaze estimation.}
Temporal information has proven effective for 3D gaze estimation. 
Previous research developed models to learn from various inputs, 
including face, eyes, and facial landmarks using a multi-stream recurrent CNN~\cite{PalmeroSBE18}; eyes and visual stimuli using convolutional RNNs~\cite{Park2020ECCV}; raw RGB frames from in-the-wild settings using convolutional bidirectional LSTMs \cite{gaze360_2019}; and the temporal coordination of gaze, head, and body orientations using LSTMs \cite{Nonaka_2022_CVPR}. 
However, the use of such methods for the gaze following task in arbitrary scenes has been underexplored. 
The only exceptions are \cite{chong2020dvisualtargetattention} who introduced a convolutional LSTM block at the bottleneck of the heatmap prediction architecture, and \cite{miao2023patch} who leveraged temporal attention over aggregated frame-level features.
However, both approaches only showed a slight improvement
compared to their static versions, 
highlighting the challenge of exploiting temporal information for this task. 
Conceptually, the methods did not model 2D gaze direction dynamics, 
and can not be extended for multi-person gaze inference.

\mypartitle{Social gaze prediction.}
%
Several research papers in the literature are dedicated to the study of looking at each other (\LAEO) and shared attention (\SA) tasks.
For \LAEO, most methods rely on processing the head crops to obtain some gaze directional information, and then combining it with 2D or inferred 3D geometric information to predict the \LAEO label \cite{Marin-Jimenez_2019_CVPR, Doosti_Chen_Vemulapalli_Jia_Zhu_Green_2021,marin21pami,
cantarini2021hhpnet}. 
A drawback of these methods is that they process pairs of persons independently. In addition, as they only process heads and do not address gaze following, they lack global image context and do not extend easily to other social tasks like \SA.
Similarly to \cite{tu2022end,Tonini_2023_ICCV}, a recent paper~\cite{guo2022mgtr} proposed an encoder-decoder transformer architecture to predict heads and the \LAEO labels, and while achieving good results, suffers from the same drawbacks as stated above for \cite{tu2022end,Tonini_2023_ICCV}.

Regarding shared attention, the first method to address it in the wild was \cite{fan2018inferring_videocoatt}, which framed the problem as 2 tasks: the binary classification of whether shared attention occurs in a frame, 
and the inference of the location of shared attention target object.
Their method combined predicted 2D gaze cones of people in the scene with a heatmap of object region proposals, while others  \cite{sumer2020attention} directly inferred shared attention from the raw image. 
Since then, several methods leveraged combining gaze following heatmap 
predictions of all people \cite{chong2020dvisualtargetattention, tu2022end} and improved performance.
%
Nevertheless, the above task formulation \cite{fan2018inferring_videocoatt} used by all papers suffers from two main issues: (i) it cannot distinguish between multiple \SA instances if they occur in the same frame; 
(ii) it does not determine which specific people are sharing attention.
Our work solves both problems by framing the task as a binary classification between pairs of people. This formulation is more natural and has the benefit of extending to other social gaze tasks.

Finally, none of the previous works performed both social gaze prediction tasks. 
Two notable and interesting exceptions are \cite{Fan_2019_ICCV} 
who addressed the inference of gaze communication  
activities (atomic and events, including \LAEO and \SA) 
using a graph-based approach with 12 dimensional tokens, 
and \cite{chang2023gaze_gp-static} who addressed dyadic communication by proposing a gaze following style 2-branch architecture processing order-dependent pairs of people. 
However, \cite{fan2018inferring_videocoatt} predicts a single gaze 'state' for each person which is problematic as it does not allow for simultaneous \LAEO and \SA. It also does not identify the other person involved in the social gaze interaction. On the other hand, inference using \cite{chang2023gaze_gp-static} is inefficient because the model needs $\frac{\Npersons!}{(\Npersons-2)!}$ forward passes to consider all pairwise relationships for a given scene with $\Npersons$ people.
Further, both methods do not address the gaze following task.

\section{Architecture}
\label{sec:architecture}

\begin{figure*}[t]
    \vspace{-2mm}
    \centering
    \includegraphics[width=\textwidth]{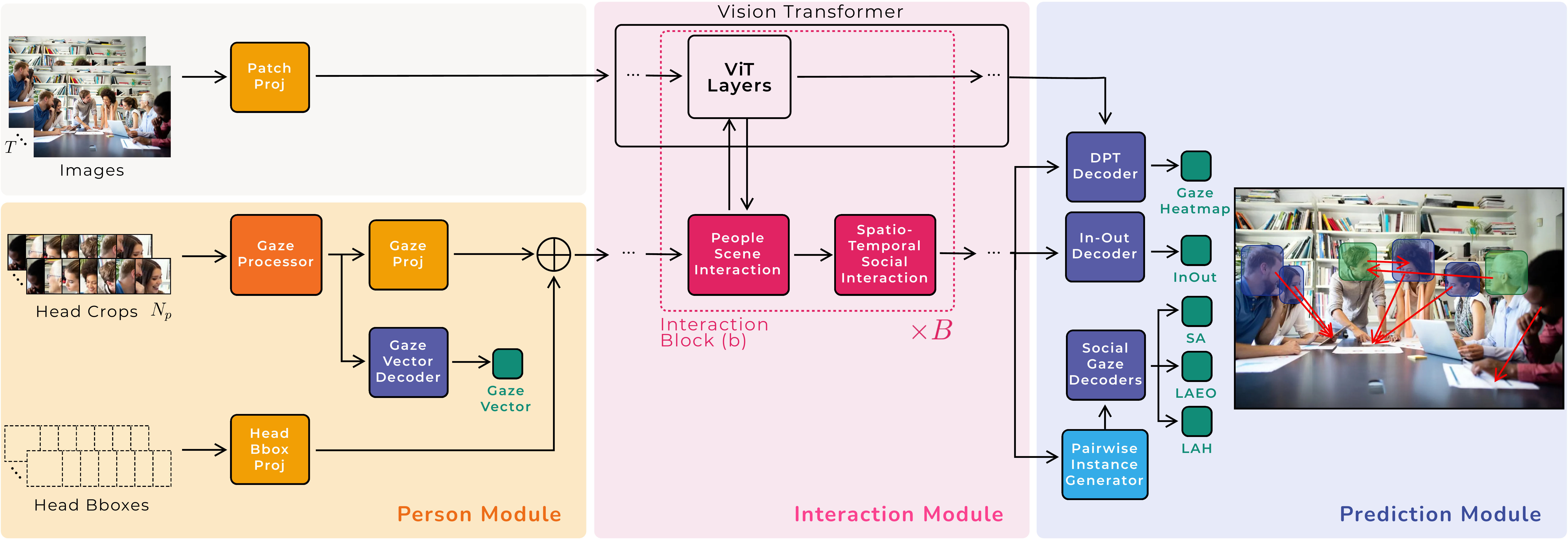}
    \vspace{-1mm}
    \caption{Proposed architecture for multi-person temporal gaze following and social gaze prediction. First, the Person Module (left) processes the set of head crops and bounding boxes to extract a sequence of person tokens for each person. In parallel, the ViT tokenizer processes the sequence of frames to extract frame tokens. Next, the Interaction Module (middle) jointly processes the person and frame tokens, iteratively updating them through people-scene interactions and spatio-temporal social interactions. Finally, the Prediction Module (right) processes the resulting frame and person tokens to infer a sequence of gaze heatmaps and in-out gaze labels for each person, as well as pair-wise social gaze labels for \LAH, \LAEO, and \SA.
    }
    \vspace{-2mm}
    \label{fig:architecture}
\end{figure*}

Our approach is described in Fig.~\ref{fig:architecture}, 
and comprises 3 different modules: the Person Module, the Interaction Module, and the
Prediction Module.
The model takes as input a sequence of $\Ntimesteps$ frames $\inputimage_{1:\Ntimesteps}$, as well as the head bounding box tracks $\{ \HeadSeqBbox_{i}=\inputheadbbox_{i,1:\Ntimesteps}, i=1,\ldots,\Npersons\}$ and corresponding head crops $\{ \HeadSeqCrop_{i}=\HeadCrop_{i,1:\Ntimesteps}, i=1,\ldots,\Npersons\}$ which are assumed to have been extracted for each of the \Npersons persons. We detail the modules in the next sections.

\subsection{Person Module}

\vspace{-1mm}

This module aims to model person-specific information relating to gaze and head location. It consists of two components.


\mypartitle{Gaze Processor.}
This component aims to capture all gaze-related information  
(direction, dynamics). 
First, individual head crops $\HeadCrop_{i,t}$ are processed by the Gaze Backbone \gazebackbone to produce gaze embeddings as
$ \gazetokenstatic_{i,t} = \gazebackbone ( \HeadCrop_{i,t} ). $
Then, to model the gaze dynamics, we rely on  a Temporal Gaze Encoder \gazetempencoder to process the sequence $\gazetokenstatic_{i,1:T}$ of gaze embeddings  of a person and obtain their temporal counterparts: 
$\gazetokentemp_{i,1:T} = \gazetempencoder ( \gazetokenstatic_{i, 1:T} ).$
\gazetempencoder is implemented as a single Transformer layer with self-attention. 
Finally, to supervise the learning of relevant gaze embeddings, 
we attach a Gaze Vector Decoder that  predicts a person's 2D gaze 
vector at each time step,
$\gazevector_{i,t} = \gazevectorMLP ( \gazetokentemp_{i, t} )$,
where \gazevectorMLP is implemented as 2-layer MLP.

\mypartitle{Person tokenization.}
Finally, the person tokens are obtained by projecting the temporal gaze embeddings and head box locations  using learnable linear layers to embeddings with the token dimension and adding them together: \\[-3mm]
\begin{equation}
    \persontoken_{i,t} = \gazeproj(\gazetokentemp_{i,t}) + \bboxproj(\inputheadbbox_{i,t}).
\end{equation}

\subsection{Interaction Module}
\vspace{-1mm}

This module aims at modeling the exchange of information between people and the scene, as well as the spatio-temporal social interactions between people.
To achieve this, we designed an architecture consisting of $B$ blocks,
each comprising a People-Scene Interaction and Spatio-Temporal Social Interaction component. 
Each block processes the set of output person tokens  $\{ \inpersontoken_{i,1:T} \}$  and frame tokens $\inimgtokens_{t}$ from the previous block, and returns updated tokens after a series of interactions through the components.
The input to the first block is the person tokens $\{ \persontoken_{i,1:T} \}$ from the Person Module, and the frame tokens $\imgtokens_{t}$ from the ViT image tokenizer.

We describe the People-Scene Interaction and Spatio-Temporal Social Interaction components below.

\mypartitle{People-Scene Interaction.}
This component models the interactions between people and the scene which 
can be useful for inferring gaze to scene objects or body parts 
like hands as well as inferring some global context. 
It is loosely inspired by~\cite{chen2022vision_vitadaptor} and proceeds 
in 3 steps: 
(i) a Person-to-Scene Encoder \persontoscene processes the frame tokens $\inimgtokens_{t}$ and frame-level person tokens $\{ \inpersontoken_{i,t}\}$ to update the frame tokens with person 
information relevant to gaze (superscript p):
\begin{equation}
    \peopleimgtokens_{t} = \persontoscene ( \inimgtokens_{t}, \{ \inpersontoken_{i,t} \} ).
\end{equation}
\persontoscene is implemented as a single Transformer layer with cross-attention, where $\inimgtokens_{t}$ generate the queries and $\{ \inpersontoken_{i,t}, i=1 \}$ generate the keys and values. 
(ii) the updated frame tokens pass through a set of ViT layers 
$\vitlayer_{b-1}:\vitlayer_{b}$ to process the scene information, 
resulting  in the output frame tokens for the block:
\begin{equation}
    \outimgtokens_{t} = \vit_{\vitlayer_{b-1}:\vitlayer_{b}} ( \peopleimgtokens_{t} ).
\end{equation}
(iii)  a Scene-to-Person Encoder \scenetoperson processes the frame-level person tokens $\{ \inpersontoken_{i,t}, i=1,\cdots,\Npersons \}$ 
and obtained frame tokens $\outimgtokens_{t}$ to update the person tokens so they may capture location information
regarding salient items looked at by people in the scene (superscript s):
\begin{equation}
    \scenepersontoken_{i,t} = \scenetoperson ( \{ \inpersontoken_{i,t} \}, \outimgtokens_{t}).
\end{equation}
\persontoscene is also implemented as a single Transformer layer with cross-attention, where the set $\{ \inpersontoken_{i,t}\}$ 
generates the queries and $\outimgtokens_{t}$ 
generates the keys and values.

\mypartitle{Spatio-temporal Social Interaction.}
This component allows the sharing of information between people 
and of the alignment of their representations for social gaze prediction, 
including by modeling their temporal evolution. 
To achieve this, a Social Encoder \socialencoder
first processes and updates the frame-level person tokens $\{ \scenepersontoken_{i,t} \}$ 
to capture interactions between people at each frame, according to:
 $   \{ \socialpersontoken_{i,t} \} = \socialencoder ( \{ \scenepersontoken_{i,t} \}) $, 
which is followed by a Temporal Person Encoder \persontempencoder 
that processes the updated person token sequences $\socialpersontoken_{i,1:T}$ of each person $i$ and updates them to capture temporal patterns 
of attention, resulting in the output person tokens for the block:
$\outpersontoken_{i,1:T} = \persontempencoder ( \socialpersontoken_{i,1:T} )$.
%
Both \socialencoder and \persontempencoder are implemented as a single Transformer layer with self-attention.


\subsection{Prediction Module}
\label{sec:prediction-module}
\vspace{-1mm}

It processes the set of output person 
$\{ \outpersontoken_{i,1:T}\}$ and
frame $\outimgtokens_{t}$ tokens from 
all Interaction Module blocks to predict the gaze heatmaps and in-out labels 
for each person, as well as the pair-wise social gaze labels.


\mypartitle{Gaze Heatmap Prediction.}
This is an important and novel aspect of our architecture. 
%
Here, we take inspiration from the DPT decoder~\cite{ranftl2021vision_dpt} 
which has achieved good results on dense prediction tasks, and we adapt it to handle multiple heatmap predictions from the same ViT outputs, 
conditioned on each person's token. 
More precisely, the standard DPT decodes the image features from multiple layers of a ViT in a Feature Pyramid Network style.  
It works by fusing at block level $b$: the 
feature maps from level $b+1$ after an upsampling stage, 
and the feature maps computed by a reassemble stage 
that processes the ViT output of block $b$~\cite{ranftl2021vision_dpt}.
We aim to apply this approach to the frame tokens 
$\{\outimgtokens_{t}, b=1:B\}$, but conditioned on 
a specific person.
In our model, this is achieved through a modification in the reassemble stage, 
in which the image feature maps produced by the standard 
reassemble stage are multiplied at every location 
(using a Hadamard product) with a projection of the person's token 
$\outpersontoken_{i,t}$ of that same block level.
This process ensures that the output gaze heatmaps are controlled by person-specific information. 
%
We provide more details in the supplementary.

\mypartitle{Social Gaze Prediction.}
This decoder processes the person tokens from all $B$ Interaction Module blocks to predict 
the social gaze label for every pair of people in every frame.
In practice, 
%
the $B+1$ tokens $\{ \persontoken^{\text{o,} 0}_{i,t} \ldots  \persontoken^{\text{o,} B}_{i,t}\}$ corresponding to a single person in a frame are linearly projected and concatenated to produce a multi-scale person token $\mspersontoken_{i,t}$.
\begin{equation}
    \mspersontoken_{i,t} = \personproj^0 ( \persontoken^{\text{o,} 0}_{i,t} ) || \ldots || \personproj^B ( \persontoken^{\text{o,} B}_{i,t} ),
\end{equation}
where $||$ denotes the concatenation operation. 
Then, to predict a social gaze label, 
these tokens are concatenated  and processed by 
the decoders for \LAH (\lahMLP) and \SA (\coattMLP):
\begin{gather}
    \lookinglabel_{i \rightarrow j,t} = \lahMLP ( \mspersontoken_{i,t} || \mspersontoken_{j,t} ) 
    \mbox{ and }
    \coattlabel_{i,j,t} = \coattMLP ( \mspersontoken_{i,t} || \mspersontoken_{j,t} ).
\end{gather}
\lahMLP and \coattMLP are implemented as 3-layer MLPs with residual connections. For \LAEO, both people $i,j$ need to be looking at each other for a positive label, and either one can be looking away for a negative label. Hence, we simply compute the \LAEO label $\lookinglabel_{i \leftrightarrow j,t}$ as $\text{min}(\lookinglabel_{i \rightarrow j,t}, \lookinglabel_{j \rightarrow i,t})$.

\mypartitle{In-Out Prediction.} 
This decoder \inoutMLP processes the multi-scale person tokens $\mspersontoken_{i,t}$ to predict at every frame 
whether people are looking inside the frame or outside the frame, as 
$    \inoutlabel_{i,t} = \inoutMLP ( \mspersontoken_{i,t} ),$
where \inoutMLP is implemented as a 5-layer MLP with residual connections.

\subsection{Losses}
The total loss \loss is a linear combination of the gaze heatmap loss $\lossheatmap$, gaze vector loss $\lossangular$, social gaze losses $\losslah, \losscoatt$ and the in-out loss $\lossio$:
\vspace{-2mm}
\begin{multline}
    \loss = \losscoeffheatmap \lossheatmap + \losscoeffangular \lossangular + \losscoefflah \losslah + \losscoeffcoatt \losscoatt + \losscoeffio \lossio
\end{multline}
\vspace{-1mm}
The loss is applied to every person at every time step. 
All losses are standard: $\lossheatmap$ is defined as the pixel-wise MSE loss between the GT and predicted heatmaps, $\lossangular$ as the cosine loss, 
and the social gaze and in-out losses as binary cross-entropy losses.
Note that since \LAEO is inferred from \LAH predictions (Sec~\ref{sec:prediction-module}), we do not have any \LAEO loss.

\section{Experiments}
\label{sec:experiments}

\vspace{-1mm}

\subsection{Datasets}
\label{sec:datasets}

\vspace{-1mm}
We perform experiments on 5 gaze following and social gaze datasets.

\mypartitle{GazeFollow~\cite{recasens2017following}.}
It is a large-scale static dataset for gaze following, featuring 122K images. Most images are annotated for a single person with their head bounding box and gaze target point. The test set contains gaze point annotations by multiple annotators. Despite lower quality images and annotations, given its rich diversity, it remains a good dataset to use for pre-training.

\mypartitle{VideoAttentionTarget (VAT)~\cite{chong2020dvisualtargetattention}.}
It is a video dataset, annotated with head bounding boxes, gaze points, and inside vs outside frame gaze for a subset of the people in the scene. It contains 1331 clips collected from 50 shows on YouTube. 

\mypartitle{ChildPlay~\cite{tafasca2023childplay}.}
It is a recent video dataset for gaze following, annotated with head bounding boxes, gaze points, and a  label indicating 7 non-overlapping gaze classes including inside frame, outside frame  and gaze shifts. It contains 401 clips from 95 YouTube videos, and features children playing and interacting with other children and adults. We further extend ChildPlay with speaking status annotations  (see supplementary for details).

\mypartitle{VideoCoAtt~\cite{fan2018inferring_videocoatt}.}
It is a video dataset for shared attention estimation, containing 380 videos or 492k frames from TV shows. When a shared attention behavior occurs (i.e. about 140k frames), the relevant frames are annotated with the bounding box of the target object, as well as the head bounding boxes of the people involved.

\mypartitle{UCO-LAEO~\cite{Marin-Jimenez_2019_CVPR}.}
UCO-LAEO is a video dataset for \LAEO estimation, annotated with head
bounding boxes, and a label  indicating whether two heads are \LAEO. It contains 22,398 frames from 4 TV shows.

Two other interesting datasets with annotations for multiple social gaze behaviours are VACATION~\cite{Fan_2019_ICCV} and GP-static~\cite{chang2023gaze_gp-static}. However, VACATION does not provide annotations for all social gaze behaviours when they occur simultaneously (which is linked to their method design, Sec~\ref{sec:RelatedWork}). On the other hand, GP-static only considers dyadic interactions and is not publicly available. 

\vspace{-4mm}

\subsection{Pre-processing}

We propose a novel pre-processing scheme to automatically obtain gaze following and social gaze annotations across the aforementioned datasets. 


\begin{table}[t]
\centering 
\scriptsize
\begin{tabular}{l|cccc}\toprule 

\bf{Dataset} & \bf{\ Gaze Points} & \bf{LAH} & \bf{LAEO} & \bf{SA} \\ 
\midrule

GazeFollow~\cite{recasens2017following} & 118k & 27k/493k & 0 & 0 \\
\midrule
VAT~\cite{chong2020dvisualtargetattention} & 109k & 74k/729k & 13k/461k & 16k/94k \\
ChildPlay~\cite{tafasca2023childplay} & 217k & 59k/682k & 7k/351k & 4k/55k \\
VideoCoAtt~\cite{fan2018inferring_videocoatt} & 367k & 290k/1551k & 0 & 400k/918k \\
UCO-LAEO~\cite{Marin-Jimenez_2019_CVPR} & 21k & 21k/36k & 10k/54k & 0 \\

\rowcolor{BackgroundColor}
\bf{VSGaze} & 714k & 444k/2998k & 30k/866k & 420k/1067k \\

\bottomrule 
\end{tabular} 

\caption{Obtained person-wise gaze point and pair-wise social gaze annotation (positive/negative) statistics for our datasets. VSGaze refers to the combined video datasets: VAT, ChildPlay, VideoCoatt and UCO-LAEO.}
\vspace{-6mm}
\label{tab:social-gaze-annotations}

\end{table}

\mypartitle{Gaze Target Point.}
For people sharing attention in  VideoCoAtt~\cite{fan2018inferring_videocoatt}, we compute their gaze points as the center of the shared gaze object's bounding box. 
Similarly, for a person pair \LAEO in UCO-LAEO~\cite{Marin-Jimenez_2019_CVPR}, we compute their gaze points as the center of the other person's head bounding box. 

\mypartitle{Head Bounding Boxes and Tracks.}
As each dataset contains annotations for only a subset of people in the scene, we detect and track all missing heads using the pre-trained Yolov5 head detection model~\cite{glenn_jocher_2022_7002879_yolov5} used by Tafasca \etal~\cite{tafasca2023childplay} and the ByteTrack tracking algorithm~\cite{zhang2022bytetrack}. We manually verified the accuracy of the obtained detections and tracks.

\mypartitle{\LAH.}
We generate \LAH annotations for all datasets. 
To do so, similarly to Tafasca \etal~\cite{tafasca2023childplay}, 
we check whether the gaze point for an annotated person falls inside any other person's head bounding box. For the GazeFollow test set, at least 2 of the annotated gaze points should fall inside another person's head bounding box. 

\mypartitle{\LAEO.}
We use the provided annotations for UCO-LAEO.
For VAT and ChildPlay, we generate \LAEO annotations by using the \LAH annotations, 
checking whether the \LAH target for a pair of people corresponds to the other person.
We cannot obtain \LAEO for GazeFollow as most images are annotated for a single person, and for VideoCoAtt because if a person is an \SA target,
they are not in the set of people sharing attention and their gaze is not annotated.  

\mypartitle{\SA.}
We use the provided annotations for VideoCoAtt. For VAT and ChildPlay,
we generate novel \SA annotations from the \LAH annotations, 
checking whether two person share their attention to the same third person.
We cannot obtain \SA for GazeFollow as most images are annotated for a single person, and for UCO-LAEO as a pair of people annotated with \LAEO cannot be sharing attention. 

Henceforth, we refer to the set of processed \textbf{V}ideo datasets with \textbf{S}ocial gaze and \textbf{Gaze} following annotations as \textbf{VSGaze}.

\mypartitle{Annotation statistics} are summarized  in Table~\ref{tab:social-gaze-annotations}. 
Overall, VideoCoatt is the largest source of annotations except for \LAEO. 
We also see that the pair-wise annotations are heavily skewed towards negative cases. The statistics further provide insight into the content of the datasets. As VAT, VideoCoAtt and UCO-LAEO contain clips from TV shows, there are many more instances of looking at other people and at each other. On the other hand for ChildPlay, \LAH mainly occurs when the supervising adult looks at a child and there is limited \LAEO. 

\vspace{-2mm}

\subsection{Implementation Details}

\vspace{-1mm}

The Gaze Backbone \gazebackbone is a ResNet18 pre-trained on the Gaze360 dataset~\cite{gaze360_2019}, and processes the head crops at a resolution of $224 \times 224$. We use a ViT-base model~\cite{arnab2021vivit} \vit initialized with MultiMAE weights~\cite{bachmann2022multimae} to process the scene at a resolution of $224 \times 224$. 
For all our experiments, we use $T=5$ frames with a temporal stride of $3$. To allow for batch training, we randomly sample up to $\Npersons=4$ people in a scene (padding in case there are less). During testing, we evaluate per sample and consider all people. 
%
%
The Interaction Module consists of $4$ blocks, interacting with \vit at layers $\vitlayer_b = \{ 2,5,8,11 \}$. The loss coefficients are set as $\losscoeffheatmap=1000, \losscoeffangular=3, \losscoeffio=2$ and $\losscoefflah=\losscoeffcoatt=1$.
To counter the class imbalance, we weight positive samples for social gaze loss by $2$.

\vspace{-1mm}

\subsection{Training and Validation}

\vspace{-1mm}

We follow standard practice~\cite{chong2020dvisualtargetattention} 
by first training the static version of our model (i.e. with no temporal attention) on GazeFollow. It is trained for 20 epochs with a learning of rate of $1 \times 10^{-4}$. 
The resulting weights then serve as initialization for our proposed temporal model.
We freeze the ViT \vit and train the temporal model on VSGaze for another 20 epochs with a learning rate of $3 \times 10^{-6}$. 
We use the AdamW optimizer~\cite{loshchilov2018decoupled_adamw} with warmup and cosine annealing. 
For validation, we use the provided splits for UCO-LAEO, VideoCoAtt and ChildPlay, and the splits proposed by Tafasca \etal~\cite{tafasca2023childplay} for GazeFollow and VAT.

\vspace{-1mm}

\subsection{Evaluation}
\label{sec:evaluation}

\vspace{-2mm}

\mypartitle{Gaze Following.}
We use the standard metrics: 
\begin{compactitem}
    \item \textit{AUC}: for GazeFollow, the predicted heatmap is compared against a binary GT map with value 1 at annotated gaze point positions, to compute the area under the ROC curve. 
    %
    \item \textit{Distance (Dist.)}: the arg max of the heatmap provides the gaze point. 
    We can then compute the L2 distance between the predicted and GT gaze point on a $1\times 1$ square. For GazeFollow, we compute Minimum (Min.) and Average (Avg.) distance against all annotations.
    \item \textit{In-Out AP (AP$_{\text{IO}}$)}: it is the Average Precision (AP) 
    of the In-Out frame gaze prediction scores.
\end{compactitem}

\mypartitle{Social Gaze.}
We use the average precision (AP) metric for evaluating social gaze prediction tasks. 
For the \LAH task, a sample is an individual person $i$, 
and at inference, it is assigned the person $\hat{j}$ 
for which the $\lookinglabel_{i \rightarrow j, t}$ 
is the largest. 
Hence, for a GT positive case, the prediction 
will be considered as a true positive if $j^* = \hat{j}$ AND the score 
$\lookinglabel_{i \rightarrow \hat{j}, t}$ is above the threshold used 
for computing the AP curve. Otherwise, the prediction is a false negative.
Similarly, a GT negative is a true negative if $\lookinglabel_{i \rightarrow \hat{j}, t}$ is below the threshold for computing the AP curve, 
otherwise it is a false positive. 

For the \LAEO and \SA tasks, a sample is a pair of people, 
and a positive case is when \LAEO and \SA occurs, and negative otherwise. 
For \LAEO, as with \LAH, we consider that person $i$ as the most likely 
eye contact with the person $j$ having the highest predicted score 
$\lookinglabel_{i \leftrightarrow j, t}$ and then set 
$\lookinglabel_{i \leftrightarrow j, t}$ to 0 $\forall j \neq \hat{j}$
before computing the performance. 

\vspace{-1mm}

 \vspace{-1mm}

\section{Results}
\label{sec:results}

 \vspace{-2mm}

\subsection{Comparison against the State-of-The Art}

 \vspace{-1mm}

We compare against recent SoTA methods addressing either 
social gaze tasks or gaze following.
In addition, for fairness and to evaluate the impact of the VSGaze dataset, we also re-trained  on this dataset the static image based models of Chong~\cite{chong2020dvisualtargetattention} (Chong$_{S}$*) and Gupta~\cite{gupta2022modular} (Gupta*), as well as the temporal model of \cite{chong2020dvisualtargetattention} (Chong$_{T}$*), 
the only temporal gaze following model with available code. 

\mypartitle{Social gaze labels for gaze following methods} are 
obtained by post-processing their predicted points. For \LAH, we check whether the predicted gaze point for a person falls inside the target person's head box. For \LAEO we check the reverse as well. We compute F1 scores for both (we threshold predictions from decoders at 0.5). For \SA, we check if the distance between two people's predicted gaze points is within a threshold and compute an AP score. 
\begin{table}[t]
\centering 
\scriptsize

\begin{tabular}{llc|ccccc}\toprule 

\bf{Dataset} & \bf{Method} & \bf{PP} & \bf{Dist. $\downarrow$} & \bf{AP$_{\text{IO}}\uparrow$} &  \bf{F1$_\text{\LAH}\uparrow$} & \bf{F1$_\text{\LAEO}\uparrow$} & \bf{AP$_\text{\SA}\uparrow$} \\ 

\midrule

\multirow{3}{*}{VAT~\cite{chong2020dvisualtargetattention}} & Chong$_{S}$*~\cite{chong2020dvisualtargetattention} & \cmark & 0.133 & 0.798 & 0.785 & 0.506 & 0.205 \\

& Chong$_{T}$*~\cite{chong2020dvisualtargetattention} & \cmark & 0.137 & 0.843 & 0.778 & 0.480 & 0.332 \\

& Gupta*~\cite{gupta2022modular} & \cmark & 0.139 & 0.795 &  0.762 & 0.516 & 0.300 \\

 \rowcolor{BackgroundColor} & Ours-noGF & \xmark & - & - & 0.766 & 0.503 & 0.435\\

 \rowcolor{BackgroundColor} & Ours-noSoc & \cmark & \underline{0.123} & \textbf{0.847} &  \underline{0.805}  & \textbf{0.559} & 0.443\\

 \rowcolor{BackgroundColor} & Ours & \xmark & \textbf{0.116} & \underline{0.845} & 0.791  & 0.526 & \textbf{0.521}\\

 \rowcolor{BackgroundColor} & Ours-PP & \cmark & \textbf{0.116} & \underline{0.845} & \textbf{0.820} & \underline{0.557} & \underline{0.498}\\

\midrule

\multirow{3}{*}{ChildPlay~\cite{tafasca2023childplay}} & Chong$_{S}$*~\cite{chong2020dvisualtargetattention} & \cmark & 0.121 & 0.973 &  0.597& 0.497 & \textbf{0.243} \\

& Chong$_{T}$*~\cite{chong2020dvisualtargetattention} & \cmark & 0.137 & 0.985 & 0.571 & 0.400 & 0.165 \\

& Gupta*~\cite{gupta2022modular} & \cmark & \underline{0.118} & 0.979 &  0.566 & 0.433 & 0.132 \\

 \rowcolor{BackgroundColor} & Ours-noGF & \xmark & - & - &  0.609 & 0.404 & 0.207 \\

 \rowcolor{BackgroundColor} & Ours-noSoc & \cmark & 0.121 & \textbf{0.994} &  0.611 & 0.402 & 0.188\\

 \rowcolor{BackgroundColor} & Ours & \xmark & \textbf{0.115} & \underline{0.993} &  \textbf{0.682} & \underline{0.422} & 0.178\\

 \rowcolor{BackgroundColor} & Ours-PP & \cmark & \textbf{0.115} & \underline{0.993} & \underline{0.645} & \textbf{0.431} & \underline{0.214}\\

\midrule

\multirow{3}{*}{VideoCoAtt~\cite{fan2018inferring_videocoatt}} & Chong$_{S}$*~\cite{chong2020dvisualtargetattention} & \cmark & 0.122 & - &  0.794 & - & 0.266 \\

& Chong$_{T}$*~\cite{chong2020dvisualtargetattention} & \cmark & 0.126 & - & 0.790  & - & 0.337 \\

& Gupta*~\cite{gupta2022modular} & \cmark & \underline{0.116} & - & 0.815  & - & 0.347 \\

 \rowcolor{BackgroundColor} & Ours-noGF & \xmark & - & -  & 0.733 & - & \underline{0.524}\\

 \rowcolor{BackgroundColor} & Ours-noSoc & \cmark & \textbf{0.111} & - &  \underline{0.819} & - & 0.334\\

 \rowcolor{BackgroundColor} & Ours & \xmark & \textbf{0.111} & -  & 0.804 & - & \textbf{0.603}\\

 \rowcolor{BackgroundColor} & Ours-PP & \cmark & \textbf{0.111} & -  & \textbf{0.820} & - & 0.343\\

\midrule

\multirow{3}{*}{UCO-LAEO~\cite{Marin-Jimenez_2019_CVPR}} & Chong$_{S}$*~\cite{chong2020dvisualtargetattention} & \cmark & \underline{0.032} & -  & 0.986  & 0.811 & - \\

& Chong$_{T}$*~\cite{chong2020dvisualtargetattention} & \cmark & 0.064 & -  & 0.950 & 0.779 & - \\

& Gupta*~\cite{gupta2022modular} & \cmark & \textbf{0.030} & -  & 0.989 & 0.856 & - \\

 \rowcolor{BackgroundColor} &  Ours-noGF & \xmark & - & - & 0.989  & \textbf{0.939} & -\\

 \rowcolor{BackgroundColor} & Ours-noSoc & \cmark & 0.043 & -  & 0.978  & 0.842 & -\\

 \rowcolor{BackgroundColor} & Ours & \xmark & \underline{0.032} & -  & \underline{0.990}  & \underline{0.889} & -\\

 \rowcolor{BackgroundColor} & Ours-PP & \cmark & \underline{0.032} & -  & \textbf{0.995} & 0.873 & -\\

 \midrule

\multirow{3}{*}{\textbf{VSGaze}} & Chong$_{S}$*~\cite{chong2020dvisualtargetattention} & \cmark & 0.121 & 0.918  & 0.779 & 0.579 & 0.289\\

& Chong$_{T}$*~\cite{chong2020dvisualtargetattention} & \cmark & 0.130 & \textbf{0.956} & 0.764 &  0.529 & 0.331 \\

& Gupta*~\cite{gupta2022modular} & \cmark & 0.119 & 0.929 & 0.782 & 0.591 & 0.335 \\

 \rowcolor{BackgroundColor} &  Ours-noGF & \xmark & - & -  & 0.738 & 0.579 & \underline{0.515}\\

 \rowcolor{BackgroundColor} &  Ours-noSoc & \cmark & \underline{0.114} & \underline{0.945} & \underline{0.798} & \underline{0.598} & 0.339\\

 \rowcolor{BackgroundColor} &  Ours & \xmark & \textbf{0.111} & 0.940 & 0.794 & 0.589 & \textbf{0.577}\\

  \rowcolor{BackgroundColor} & Ours-PP & \cmark & \textbf{0.111} & 0.940 & \textbf{0.807} & \textbf{0.611} & 0.351\\

\bottomrule 
\end{tabular} 

\caption{Comparison against gaze following methods on VSGaze and its component datasets: VAT~\cite{chong2020dvisualtargetattention}, ChildPlay~\cite{tafasca2023childplay}, VideoCoAtt~\cite{fan2018inferring_videocoatt} and UCO-LAEO~\cite{Marin-Jimenez_2019_CVPR}. All models were trained on VSGaze. PP indicates social gaze predictions from post-processing gaze following outputs (\cmark) vs predictions from decoders (\xmark). Best results are in bold, second best results are underlined.}

\label{tab:results-VSGaze}
\vspace{-5mm}
\end{table}

\mypartitle{VSGaze.}
We first analyse the results on VSGaze which are given  in Table~\ref{tab:results-VSGaze}. Note that regarding our approach, 
for social gaze, we compute the scores by leveraging either the predictions from the respective task decoders (Ours), 
or by post-processing the gaze following outputs of 
our model (Ours-PP). 

Compared to the baselines, we observe that our model achieves 
the best performance for all tasks except for in vs out 
of frame gaze prediction. 
In particular, we achieve significant gains in the distance and  AP$_\text{\SA}$ metrics when leveraging the predicted outputs from the \SA decoder. The latter highlights the importance of modeling \SA as a classification task compared to post-processing gaze following outputs, which struggles to capture whether the gaze points for a pair of people falls on the same semantic item.

We also observe that performance trends on individual datasets can differ from the aggregated results on VSGaze. For instance, although we have a small improvement for \LAH compared to the baselines across VSGaze, we perform significantly better on ChildPlay. 

In addition, we note that better gaze following performance does not always translate to better social gaze performance. For instance, although Gupta* has a better distance score compared to Chong$_{S}$* on ChildPlay, it performs much worse for all social gaze tasks. 
This suggests the benefit of considering social gaze metrics for better characterizing the performance of gaze following models, especially its semantic gaze following performance.

\begin{table}[t]
    \centering
    \scriptsize
\begin{subtable}[c]{0.52\linewidth}
    \centering
    \begin{tabular}{l c | c c c c}
        \toprule
        \textbf{Method} & \textbf{Multi} & \textbf{AUC}$\uparrow$ & \textbf{Avg.Dist.}$\downarrow$ & \textbf{Min.Dist.}$\downarrow$ \\
        \midrule
        Fang~\cite{Fang_2021_CVPR_DAM} & \xmark & 0.922 & 0.124 & 0.067\\
        
        Tonini~\cite{tonini2022multimodal} & \xmark & 0.927 & 0.141 & -\\
        
        Jin~\cite{jin2022depth} & \xmark & 0.920 & 0.118 & 0.063 \\
        Bao~\cite{bao2022escnet} & \xmark & 0.928 & 0.122 & - \\
        Hu~\cite{hu2022gaze_object} & \xmark & 0.923 & 0.128 & 0.069 \\
        Tafasca~\cite{tafasca2023childplay} & \xmark & 0.936 & 0.125 & 0.064\\
        Chong$_{S}$~\cite{chong2020dvisualtargetattention} & \xmark & 0.921 & 0.137 & 0.077\\
        Gupta~\cite{gupta2022modular} & \xmark & 0.933 & 0.134 & 0.071\\
        
        Jin~\cite{jin2021multi} & \cmark & 0.919 & 0.126 & 0.076\\
        
        \rowcolor{BackgroundColor}
        Ours-static & \cmark & 0.929 & \underline{\textbf{0.118}} & \underline{\textbf{0.062}} \\

        \bottomrule
    \end{tabular}
    \caption{Results on GazeFollow~\cite{recasens2017following}.}
    \label{tab:results-gazefollow}
\end{subtable}
\hspace{\fill}
\begin{subtable}[c]{0.46\linewidth}
    \centering
    \begin{tabular}{l c | c c}
        \toprule
       \textbf{Method}  & \textbf{Multi} &  \textbf{Dist.$\downarrow$} & \textbf{AP$_{\text{IO}}\uparrow$}\\
       
       \midrule
       
       Fang~\cite{Fang_2021_CVPR_DAM} & \xmark & \underline{0.108} & 0.896\\
       Tonini~\cite{tonini2022multimodal} & \xmark & 0.129 & - \\
       Jin~\cite{jin2022depth} & \xmark & 0.109 & \underline{0.897}\\
       Bao~\cite{bao2022escnet} & \xmark & 0.120 & 0.669\\
       Hu~\cite{hu2022gaze_object} & \xmark & 0.118 & 0.881\\
       Tafasca~\cite{tafasca2023childplay} & \xmark & 0.109 & 0.834\\
       Chong$_{T}$~\cite{chong2020dvisualtargetattention} & \xmark & 0.134 & 0.853\\
       Gupta~\cite{gupta2022modular} & \xmark & 0.134 & 0.864\\
       
       Jin~\cite{jin2021multi} & \cmark & 0.134 & \textbf{0.880}\\

       \rowcolor{BackgroundColor}
       Ours &  \cmark & \textbf{0.110} & 0.869 \\
       
       \rowcolor{BackgroundColor}
       Ours\dag & \cmark & 0.111 & 0.869 \\
       \bottomrule
    \end{tabular}
    \caption{Results on VAT~\cite{chong2020dvisualtargetattention}.}
    \label{tab:results-vat}
    \vspace{-2mm}
\end{subtable}
\begin{subtable}[c]{0.52\linewidth}
    \centering
    \begin{tabular}{l c | c c}
        \toprule
       \textbf{Method}  & \textbf{Multi} &  \textbf{Dist.$\downarrow$} & \textbf{AP$_{\text{IO}}\uparrow$}\\
       
       \midrule
       
       Tafasca~\cite{tafasca2023childplay} & \xmark & \underline{0.107} & 0.986 \\
       Gupta~\cite{gupta2022modular} & \xmark & 0.113 & 0.983 \\
       
       \rowcolor{BackgroundColor}
       Ours & \cmark & 0.118 & \underline{\textbf{0.994}} \\

       \rowcolor{BackgroundColor}
       Ours\dag & \cmark & \textbf{0.113} & 0.993 \\
       \bottomrule
    \end{tabular}
    \caption{Results on ChildPlay~\cite{tafasca2023childplay}.}
    \label{tab:results-childplay}
\end{subtable}
\hspace{\fill}
\begin{subtable}[c]{0.46\linewidth}
    \centering
    \begin{tabular}{l | c c} 
        \toprule
       \textbf{Method}  &  \textbf{Dist.$\downarrow$} & \textbf{AP$_\text{\LAEO}\uparrow$}\\
       \midrule
       Jiminez~\cite{Marin-Jimenez_2019_CVPR} & - & 0.795\\
       Doosti~\cite{Doosti_Chen_Vemulapalli_Jia_Zhu_Green_2021} & - & 0.762\\
       Jiminez~\cite{marin21pami} & - & 0.867 \\
       
       \rowcolor{BackgroundColor}
       Ours & 0.026 & 0.963  \\

       \rowcolor{BackgroundColor}
       Ours\dag & \underline{\textbf{0.024}} & \underline{\textbf{0.974}}  \\
       \bottomrule
    \end{tabular}
    \caption{Results on UCO-LAEO~\cite{Marin-Jimenez_2019_CVPR}.}
    \label{tab:results-laeo}
\end{subtable}

\vspace{-2mm}

\caption{Comparison against task specific methods fine-tuned on individual datasets. Best multi-person results are in bold, overall best results are underlined. Multi indicates multi-person (\cmark) vs single-person (\xmark) gaze following methods. Ours is initialized from training on GazeFollow, while Ours$\dag$ is initialized from training on VSGaze.}
\label{tab:results-FT}
\vspace{-7mm}
\end{table}

\mypartitle{State-of-the-art comparison: fine tuning on individual datasets.}
Table~\ref{tab:results-FT} compares our model against task specific methods. 
For GazeFollow, we use our static model (Ours-static) that was 
trained on GazeFollow and used to initialize our model trained on VSGaze.
For the video datasets, as SoTA methods were trained (or finetuned) 
on individual datasets, for fairness we also fine-tune our model on these datasets, investigating two  initialization alternatives:  
either from the model trained on GazeFollow (Ours), 
or from the model trained on VSGaze (Ours$\dag$). 
Note that we are unable to compare against previous results for VideoCoatt due to our new pair-wise evaluation protocol that better captures \SA performance (Sec~\ref{sec:evaluation}). 

On GazeFollow and VAT, our model outperforms the only other comparable multi-person gaze following model of Jin~\cite{jin2021multi}. It also achieves competitive or better results to single-person methods, 
even those leveraging auxiliary modalities such as depth~\cite{Fang_2021_CVPR_DAM, tonini2022multimodal, bao2022escnet, jin2022depth, hu2022gaze_object, tafasca2023childplay}. 
Importantly, on the social LAEO task, we set the new state of the art on UCO-LAEO, far outperforming methods designed specifically for LAEO~\cite{Marin-Jimenez_2019_CVPR, marin21pami, Doosti_Chen_Vemulapalli_Jia_Zhu_Green_2021}.


We also note that fine-tuning using the VSGaze model initialization can improve results compared to the standard protocol of fine-tuning after training on GazeFollow (ex. distance on ChildPlay and AP$_{\text{\LAEO}}$ on UCO-LAEO). 
This suggests that training on VSGaze can leverage the complementary knowledge provided by the different tasks and datasets, 
which follows observations made in other works addressing 
multi-task training~\cite{ci2023unihcp}.

\vspace{-1mm}

\subsection{Ablations}

\vspace{-2mm}


\mypartitle{Impact of Architecture.}
Comparing the performance of our model with no social gaze losses (Ours-noSoc) against the baselines (Table~\ref{tab:results-VSGaze}), we see that it already performs on par or better than them while being much more efficient as it processes the image only once for all people in the scene. It also serves as a strong gaze following baseline to compare performance against.

\mypartitle{Impact of Social Gaze Loss.}
Our architecture can further benefit from the social gaze losses, 
showing improved gaze following performance and social gaze prediction (Ours and Ours-PP, Table~\ref{tab:results-VSGaze}). 
In particular, we observe significant gains for the shared attention compared to Ours-noSoc. 
Interestingly, the addition of the social gaze losses also better aligns the gaze following outputs for social gaze prediction. Comparing Ours-PP and Ours-noSocial, we see that performance for all social gaze tasks is improved.

\mypartitle{Impact of Gaze Following Loss.}
We additionally train our model without the standard gaze following losses: heatmap, gaze vector and in-out (Ours-noGF, Table~\ref{tab:results-VSGaze}). Across VSGaze, we see that performance for all social gaze tasks drops, which indicates that the gaze following and social gaze losses provide complementary information, and using both can give improved performance.

\mypartitle{Impact of VSGaze.}
When comparing the performance of the models trained on VSGaze (Table~\ref{tab:results-VSGaze}) against their versions fine-tuned on individual datasets (Table~\ref{tab:results-FT}), we see that the fine-tuned models always perform better. This is because fine-tuning allows the models to learn dataset specific priors (ex. more \LAH cases in VAT, Table~\ref{tab:social-gaze-annotations}). This highlights the challenge in leveraging multiple datasets: while we may expect better performance by having more data, the different priors and statistics bring additional difficulties. Nevertheless, our model trained on VSGaze is able to achieve strong performance across all datasets.

\mypartitle{Additional experiments, ablations.}
We provide additional experiments on incorporating auxiliary information in our model in the form of people's speaking status, ablations on temporal window length, as well as qualitative analysis in the supplementary. 

\vspace{-2mm}

\vspace{-1mm}

\section{Conclusion}

\vspace{-2mm}

We propose a new framework for multi-person, temporal gaze following and social gaze prediction comprising of a novel architecture and dataset. 
Through a series of experiments, we show that our model can effectively learn from a mix of video-based datasets with different statistics, to perform gaze following and social gaze prediction without sacrificing performance on any of them.
%
The trained model can then be further fine-tuned on individual datasets to improve performance towards a specific scenario or task.
%
%

We hope that our proposed framework opens new directions for modeling people's gaze behavior, and are keen to see applications of the proposed dataset and social gaze losses in other methods.
In the future, we intend to further investigate the benefits of temporal information and other auxiliary signals, including new ways of incorporating them into the architecture. We also plan to expand the VSGaze dataset to include more samples and annotations.

\bibliographystyle{splncs04}
\bibliography{main}

\clearpage
\begin{center}
    \Large
    \textbf{A Novel Framework for Multi-Person Temporal Gaze Following and Social Gaze Prediction}\\
    \vspace{0.5em}Supplementary Material \\
\end{center}

\section{Introduction}

We provide ablations on temporal window length and novel architecture components (Section~\ref{sec:more-ablations}), qualitative analysis of predictions (Section~\ref{sec:qualitative}), experiments on incorporating auxiliary information in the form of people's speaking status (Section~\ref{sec:auxiliary-info}), and details on our DPT based gaze heatmap decoder (Section~\ref{sec:dpt}).

Given that post-processing gaze following outputs for \LAH and \LAEO gives slightly better performance than using the predictions from their respective decoders (Table~\ref{tab:results-VSGaze}, Ours-PP), in all the results and visualizations below, \LAH and \LAEO predictions are obtained using the post-processing strategy, and \SA predictions are obtained from the corresponding decoder.


\section{More Ablations}
\label{sec:more-ablations}

\subsection{Temporal Window Length}

We compare performance of our model for different temporal window lengths on VSGaze in Table~\ref{tab:ablation-time}. Note that $T=1$ corresponds to a static model. We observe that incorporating temporal information can improve performance, especially in the case of shared attention. For the other metrics performance remains comparable. As a temporal window of 9 does not necessarily give better performance than a temporal window of 5, we use $T=5$ for our experiments.

We note that these observations are in contrast to those from the static (Chong$_S$) and temporal models (Chong$_T$) of~\cite{chong2020dvisualtargetattention}. As seen in Table~\ref{tab:results-VSGaze}, Chong$_{T}$ often performs worse than Chong$_S$, with especially lower scores for the distance and \LAEO metrics. This follows prior observations from the state of the art regarding temporal modelling for gaze following (Section~\ref{sec:RelatedWork}), and illustrates the challenge in leveraging temporal information.

While architecture design may be a reason for the lack of greater improvement in performance, the data itself is an important factor. Firstly, despite the larger number of samples in VSGaze compared to standard video based gaze datasets, there is high redundancy between frames so data diversity is not comparable to that of GazeFollow. Secondly, the moments where temporal information is important, such as during gaze shifts, only form a small percentage of total instances. Hence, improvements for predictions in these moments are not reflected in overall metrics. For instance, gaze shifts form less than 10\% of total instances in ChildPlay~\cite{tafasca2023childplay}. However, in our qualitative analysis (Section~\ref{sec:qualitative}) we can see situations where temporal information helps. In future work we plan to investigate new metrics for evaluating the performance of temporal models.

\begin{table}[t]
\centering 
\scriptsize

\begin{tabular}{l | ccccc}\toprule 

 \bf{$\pmb{T}$} &
 \bf{Dist. $\downarrow$} & \bf{AP$_{\text{IO}}\uparrow$} &  \bf{F1$_\text{\LAH}\uparrow$} & \bf{F1$_\text{\LAEO}\uparrow$} & \bf{AP$_\text{\SA}\uparrow$} \\ 

\midrule

1 & \underline{0.111} & \textbf{0.946} & 0.804 & \underline{0.608} & 0.555\\
\rowcolor{BackgroundColor} 5 & \underline{0.111} & 0.940 & \underline{0.807} & \textbf{0.611} & \textbf{0.577}\\
9 & \textbf{0.109} & \underline{0.943} & \textbf{0.808} & 0.599 & \underline{0.563}\\

\bottomrule

\end{tabular}
\caption{Ablations for different temporal window lengths $T$ on VSGaze. Best results are in bold, second best results are underlined.}
\label{tab:ablation-time}
\vspace{-2mm}
\end{table}

\subsection{Architecture Components}

\begin{table}[t]
\centering 
\scriptsize

\begin{tabular}{l | ccccc}\toprule 

 \bf{Method} &
 \bf{Dist. $\downarrow$} & \bf{AP$_{\text{IO}}\uparrow$} &  \bf{F1$_\text{\LAH}\uparrow$} & \bf{F1$_\text{\LAEO}\uparrow$} & \bf{AP$_\text{\SA}\uparrow$} \\ 

\midrule

Ours-noI$_{ppt}$ & \underline{0.111} & 0.937 & 0.801 & \underline{0.606} & \underline{0.547}\\
Ours-noI$_{sp}$ & \textbf{0.107} & 0.936 & \textbf{0.810} & 0.585 & 0.545\\
Ours-noI$_{ps}$ & 0.115 & 0.939 & 0.795 & 0.604 & 0.538\\
Ours-noDPT & 0.114 & \textbf{0.941} & 0.799 & 0.587 & 0.528\\
\rowcolor{BackgroundColor} Ours & \underline{0.111} & \underline{0.940} & \underline{0.807} & \textbf{0.611} & \textbf{0.577}\\

\bottomrule

\end{tabular}
\caption{Ablations on different novel components of our architecture on VSGaze. I$_{ppt}$ refers to the Spatio-Temporal Social Interaction component, I$_{sp}$ refers to the Scene-to-Person encoder, I$_{ps}$ refers to the Person-to-Scene encoder and DPT refers to the gaze heatmap decoder. Best results are in bold, second best results are underlined.}
\label{tab:ablation-architecture}
\vspace{-5mm}
\end{table}

We systematically remove different novel components of our architecture and provide results in Table~\ref{tab:ablation-architecture}.  

\mypartitle{Interaction Module.}
Removing the Person-to-Scene Interaction encoder \persontoscene (Ours-noI$_{ps}$) has the largest impact on performance, especially for distance, \LAH and \SA. Without this encoder, the scene tokens cannot access the person tokens, so encoding their gaze relevant salient items is much harder.
Removing the Scene-to-Person Interaction encoder \scenetoperson (Ours-noI$_{sp}$) actually improves the distance score, but decreases \LAEO and \SA performance. Without this encoder, the person tokens cannot access the scene tokens, so they cannot capture the locations of gazed at salient items. As the scene tokens may be able to adapt accordingly, gaze following performance is not impacted negatively.
Finally, removing the Spatio-Temporal Social Interaction component \socialencoder, \persontempencoder (Ours-noI$_{ppt}$) decreases \SA performance. Without this component, there is no interaction between person tokens, so identification of social dynamics is hindered.

\mypartitle{DPT Decoder.}
We replace our proposed modified DPT decoder for gaze heatmap prediction (Section~\ref{sec:dpt}) with a simpler decoder (Ours-noDPT). This decoder projects the scene and person tokens from the last Interaction block, performs a dot product between them, and then upscales the output to the heatmap resolution. Using the simpler decoder results in drops in performance for the distance, \LAEO and \SA metrics. Unlike the DPT, it lacks multi-scale representations which impacts heatmap prediction and supervision of tokens.

Overall, we observe that removing the components tends to impact shared attention performance. This is similar to the observation in the previous section regarding temporal information. Unlike \LAH and \LAEO where the target is always another person, \SA is more challenging as the shared attention target can be any person or object/point. Hence, this task may be benefitting more from additional information or architecture components. 

\section{Qualitative Analysis}
\label{sec:qualitative}

\subsection{Qualitative Results and Comparisons}

\begin{figure}[t]
    \centering
    \includegraphics[width=0.6\columnwidth]{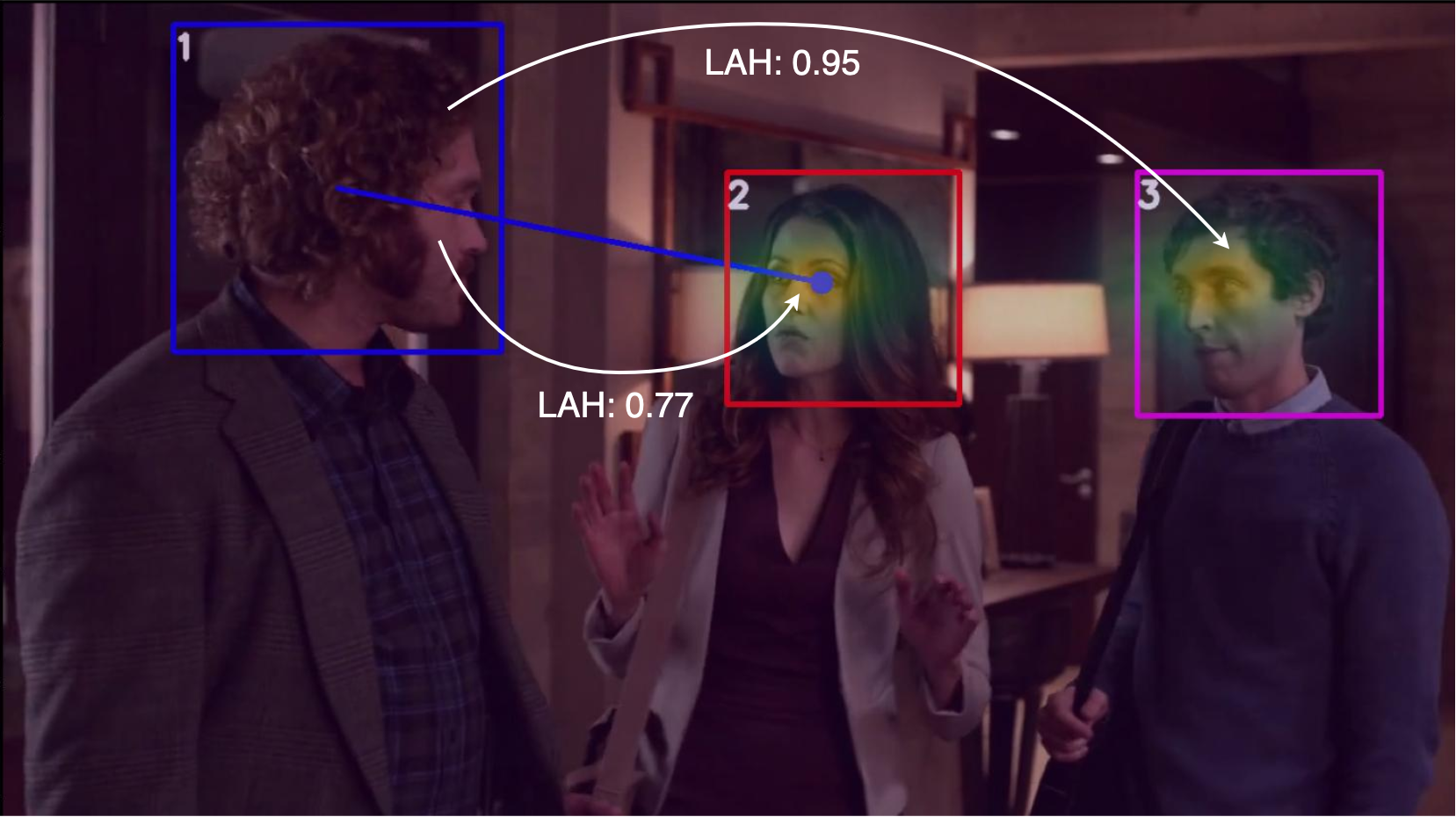}
    \caption{An illustration of the few cases where the predicted gaze point does not match with the predicted \LAH label.
    The uncertainty in the gaze target is reflected in the heatmap, while the uncertainty in the LAH target is reflected in the LAH scores.
    }
    \label{fig:confusion}
    \vspace{-2mm}
\end{figure}

\begin{figure}[t]
    \centering
    \begin{subfigure}{\textwidth}
        \includegraphics[width=\textwidth]{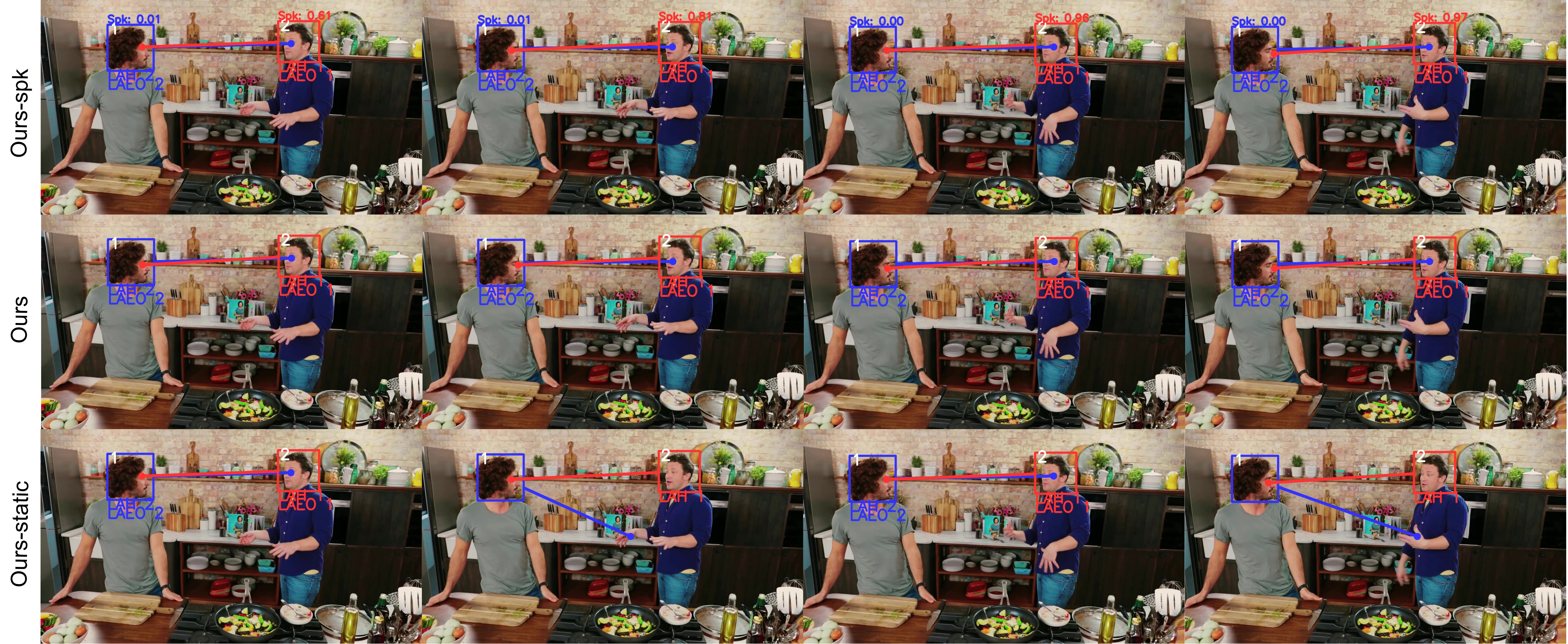}
        \caption{Ours-static fails to recognise person 1 blinking in frames 2,4}
    \end{subfigure}
    \begin{subfigure}{\textwidth}
        \includegraphics[width=\textwidth]{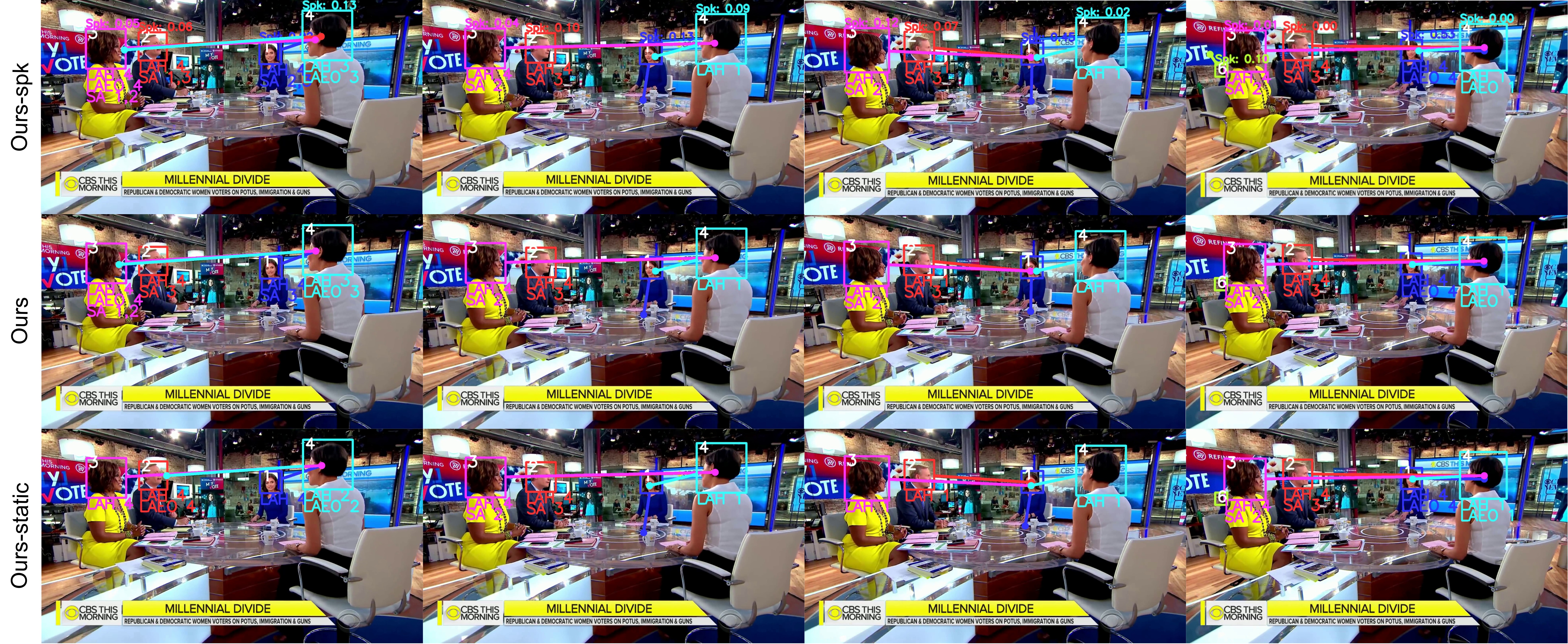}
        \caption{Ours-static misses shared attention behaviour in frames 1,3}
    \end{subfigure}
    \begin{subfigure}{\textwidth}
        \includegraphics[width=\textwidth]{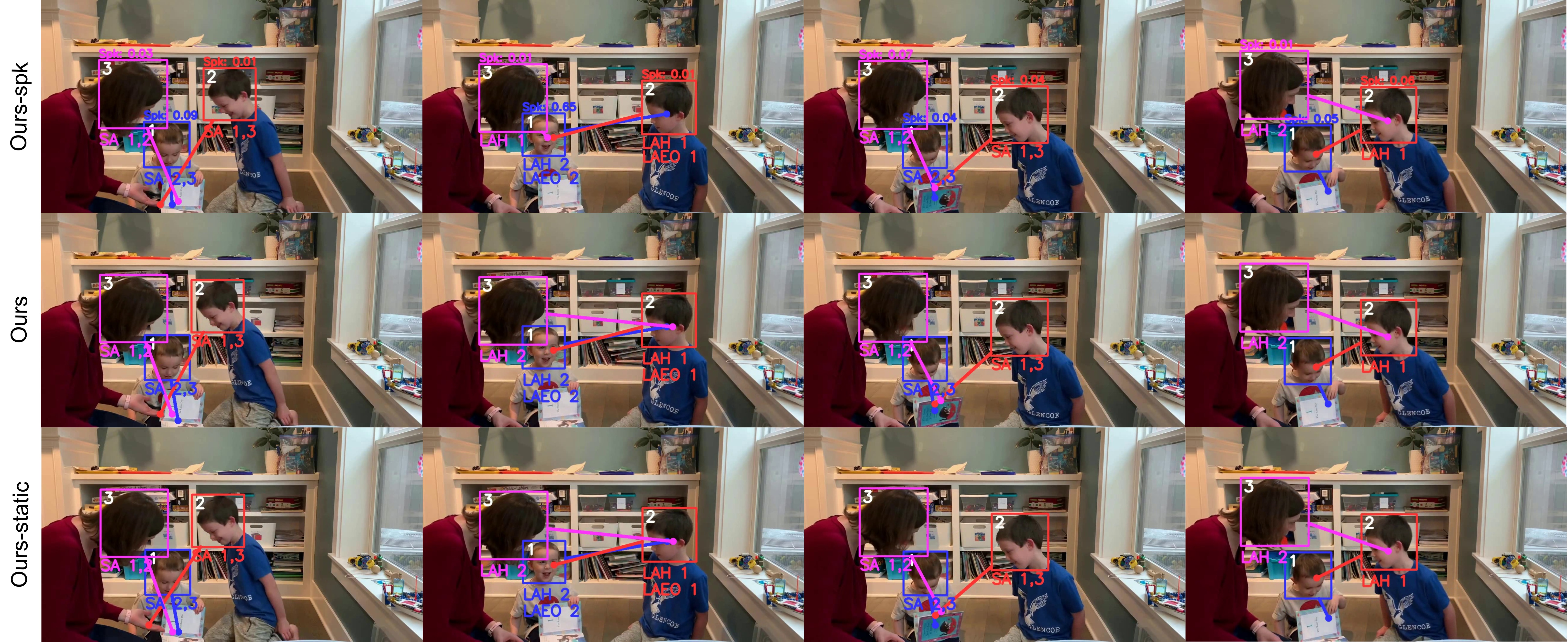}
        \caption{Ours-spk captures the right target for person 3 in frame 2}
    \end{subfigure}
    \caption{Qualitative results of our proposed model (Ours), our model with speaking information (Ours-spk) and our model without temporal information (Ours-static). When the target is predicted to be inside the frame, we display the predicted gaze point and the social gaze tasks with the associated person id(s).}
    \label{fig:qualitative-comparison}
\end{figure}

We provide qualitative results from our models in Figure~\ref{fig:qualitative-comparison}. We observe that all models perform well, accurately capturing people's gaze targets and social gaze behaviour. We also see that incorporating temporal and speaking information can help improve predictions. 

In the first sequence, the static model occasionally makes an error for person 1, and picks person 2's hands as the predicted gaze target. This is because it cannot distinguish blinking from when a person lowers their gaze. On the other hand, the temporal models recognize blinking and maintain the target as person 2's face. In the second sequence, the static model misses shared attention between persons 1,2,3 in the first frame and persons 2,3 in the third frame. This sequence is challenging due to the presence of subtle head motions which the temporal models can better capture. In the third sequence, both the static model and our proposed model make an error when predicting person 3's gaze target in frame 2. However, our model with speaking information recognizes person 1 as the target given their high speaking score.
\vspace{-2mm}

\subsection{Alignment Between Gaze Following Outputs and Social Gaze Decoders}

We perform analysis to understand the difference in performance between post-processing gaze following outputs or using the predictions from the task specific decoders. For \LAH, we find that predictions for the two schemes align 87\% of the time. On visualizing outputs, we identify that cases where they don't match are usually where the model is confused between two potential targets. So the arg max of the gaze heatmap for obtaining the gaze point picks one target, while the arg max of the \LAH scores picks the other target. This confusion in target selection is illustrated for person 1 in Figure~\ref{fig:confusion}. We see that the predicted gaze heatmap highlights both person 2 and 3, and similarly, the LAH scores for looking at both persons 2 and 3 are high.

\section{Incorporating Auxiliary Information}
\label{sec:auxiliary-info}

Previous studies on analyzing conversations during meetings have shown that people usually look at the other speaking participants \cite{StiefelhagenFocusMultiCues2002},
and such cues can be exploited for gaze target selection \cite{Otsuka:ICMI:2005}. Hence, we expect that identifying speaking persons can provide better scene understanding for gaze following, and help recognize attentiveness towards people, especially speakers. The latter is especially important in  autism diagnosis, 
as eye contact is closely monitored by the clinician when they
call out to the tested child~\cite{dsm5}. 

\subsection{ChildPlay-audio}
\label{sec:childplay-audio}

Given the importance of having joint speaking and gaze information 
especially regarding children, we extended the ChildPlay gaze dataset~\cite{tafasca2023childplay} with speaking status annotations. 

\mypartitle{Annotation Protocol.}
We mark the speaking status for every gaze annotated person in ChildPlay. It is defined by a set of 5 non-overlapping labels:
\begin{compactitem}
    \item \textit{Speaking}: the person is very likely speaking;
    \item \textit{Not-speaking}: the person is very likely not speaking;
    \item \textit{Vocalizing}: the person is very likely making a sound with their mouth (ex. to draw attention);
    \item \textit{Laughing}: the person is very likely laughing (this does not include smiling);
    \item \textit{Not-annotated}: the status cannot be inferred.
\end{compactitem}
Note that in some sequences, the person's face may be occluded or the audio may not matching with the video (ex. in the case of voice-overs). 
Nevertheless, it is still often possible to estimate the speaking status based on  head dynamics and and gaze (of others), similar to when annotating for gaze following. Hence, we ask annotators to mark the speaking status 
when it very likely a particular label, otherwise 
the 'Not-annotated' label has to be selected. 

\begin{figure}[t]
    \centering
    
\end{figure}

\mypartitle{Annotation Statistics.}
They are given in Figure~\ref{fig:childplay-audio-stats}. We observe that children mostly do not speak, while adults are more balanced between speaking and not speaking. This makes sense as  ChildPlay 
clips mainly contain children focused on their play activities, with adults supervising them. 
We also compute the percentage of cases when a speaking person\footnote{Here onward, 'speaking' includes the speaking, laughing and vocalizing labels.} is looked at by any of the annotated people. 
The statistics indicate that both children and adults are more likely to be looked at when speaking. 
However, it is worth noting that adults are not looked at much overall, as clips mostly contain a single adult and the children are focused on their play activities. This is in contrast to meeting situations where the speaking person is looked at most of the time~\cite{StiefelhagenFocusMultiCues2002}.

\begin{table}[t]
    \centering
    \begin{minipage}{0.48\linewidth}
        \includegraphics[width=\columnwidth]{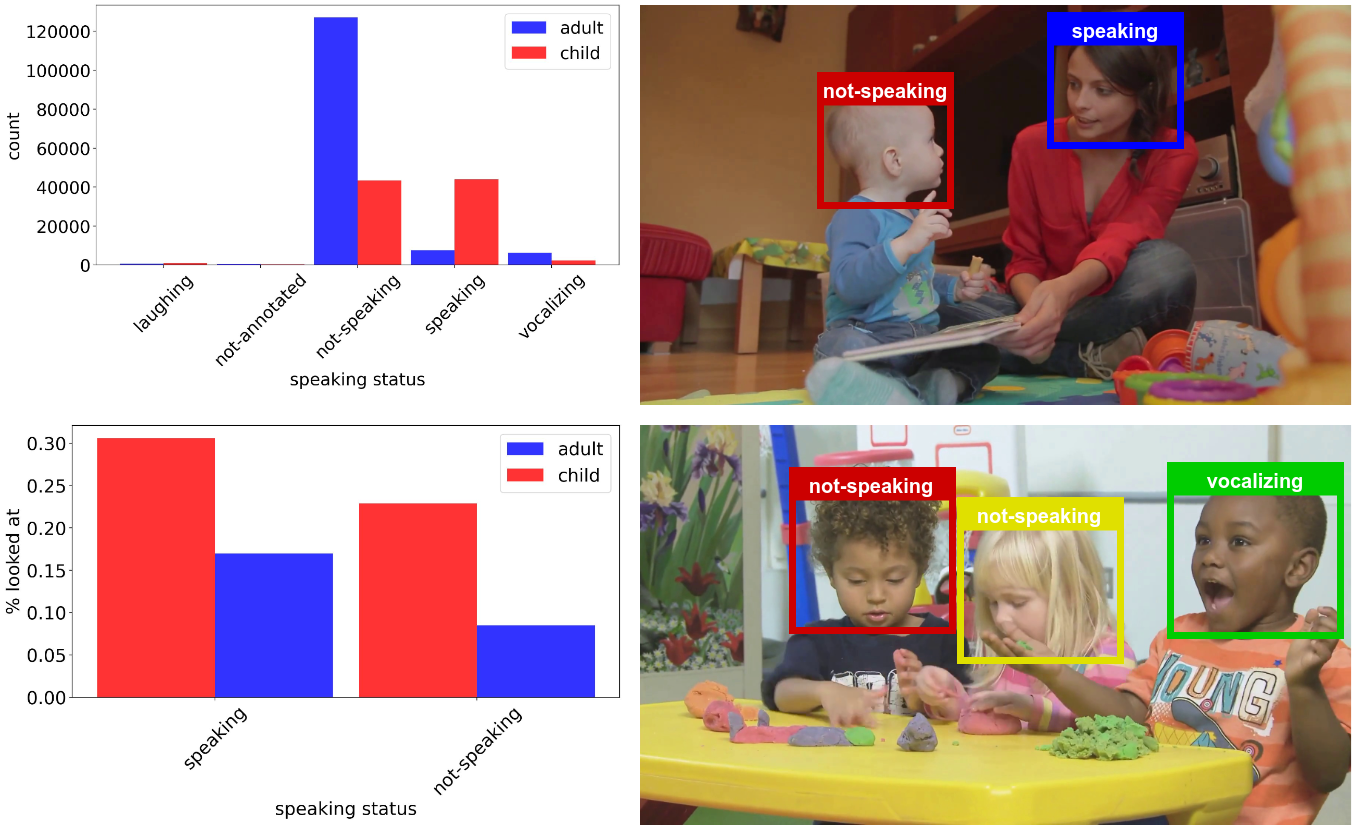}
        \captionof{figure}{Annotation statistics and samples for ChildPlay-audio.}
        \label{fig:childplay-audio-stats}
    \end{minipage}
    \hspace{\fill}
    \begin{minipage}{0.48\linewidth}
    \scriptsize
    \begin{tabular}{L{1.5cm} | M{2.2cm} M{1.8cm}}
        \toprule
         \textbf{Method} &  \textbf{AVA-activespeaker}~\cite{roth2020ava_activespeaker} & \textbf{ChildPlay-audio}\\
         \midrule 
       random  & 25.06 & 26.66 \\
       SPELL~\cite{min2022learning_spell} & 85.30 & 56.81 \\
       \bottomrule
    \end{tabular}
    \vspace{2mm}
    \caption{mAP for active speaker detection on AVA-activespeaker and ChildPlay-audio using a SoTA method.}
    \label{tab:speaking-prediction}
    \end{minipage}
    \vspace{-2mm}
\end{table}

\mypartitle{Speaking Status Prediction.}
We re-trained a state of the art model for active speaker detection~\cite{min2022learning_spell} using the visual-only modality on the large-scale AVA-activespeaker benchmark~\cite{roth2020ava_activespeaker}. 
Table~\ref{tab:speaking-prediction} provides prediction results of the model on the AVA and ChildPlay test sets.
We see that while the model performs well on AVA, it does not generalize well to ChildPlay, which indicates the challenge of detecting speech in natural scenes with children, 
and the potential of ChildPlay-audio as a challenging new benchmark for speaking status prediction, as it contains significantly different settings in terms of people (children), poses (sitting) and environments (schools, daycare centers).
\vspace{-2mm}

\subsection{Experiments and Results}

\begin{table}[t]
\centering 
\scriptsize

\begin{tabular}{l l | ccccc}\toprule 

 \bf{Dataset} & \bf{Method} &
 \bf{Dist. $\downarrow$} & \bf{AP$_{\text{IO}}\uparrow$} &  \bf{F1$_\text{\LAH}\uparrow$} & \bf{F1$_\text{\LAEO}\uparrow$} & \bf{AP$_\text{\SA}\uparrow$} \\ 

\midrule

\multirow{3}{*}{ChildPlay} & Ours-spk & 0.114 & 0.993 & 0.650 & 0.450 & 0.309\\
& Ours-spk\ddag & \textbf{0.113} & \textbf{0.994} & \textbf{0.655} & \textbf{0.457} & \textbf{0.322}\\
& Ours & \textbf{0.113} & 0.993 & 0.651 & 0.431 & 0.318\\

\midrule

\multirow{2}{*}{VSGaze} & Ours-spk & \textbf{0.111} & 0.938 & 0.806 & 0.601 & \textbf{0.590}\\
& Ours & \textbf{0.111} & \textbf{0.940} & \textbf{0.807} & \textbf{0.611} & 0.577\\

\bottomrule

\end{tabular}
\caption{Performance for incorporating people's speaking information in our model. Best results are in bold. $\ddag$ indicates ground truth speaking information.}
\label{tab:results-speaking}
\vspace{-7mm}
\end{table}

To incorporate speaking information in our model, we adapt the Person Module (Section~\ref{sec:architecture}). Specifically, we obtain speaking scores $\speakingstatus_{i,t}$ for each person using the re-trained active speaker detection model described above, linearly project the score to the token dimension using \speakingproj, and add this token to the person token. Hence, the new person token is obtained as:
\begin{equation}
    \persontoken_{i,t} = \gazeproj(\gazetokentemp_{i,t}) + \speakingproj(\speakingstatus_{i,t}) + \bboxproj(\inputheadbbox_{i,t}).
\end{equation}
We note that this formulation can easily be extended to incorporate other kinds of person-specific auxiliary information.

We provide results for incorporating speaking information in Table~\ref{tab:results-speaking}. On VSGaze, once again we see improvements for \SA. We also provide results for our model initialized with VSGaze training and fine-tuned on ChildPlay. Including speaking information improves \LAEO performance, albeit with a slight drop in \SA performance. However, using ground truth speaking information, we see similar or better performance for all metrics. As seen in Table~\ref{tab:speaking-prediction}, our active speaker detection model performs poorly on ChildPlay, hence we may benefit from better models for this task.

\vspace{-2mm}

\section{Gaze Heatmap Prediction Details}
\label{sec:dpt}

\begin{figure}[t]
    \centering
    \begin{subfigure}{0.45\linewidth}
        \includegraphics[width=\linewidth]{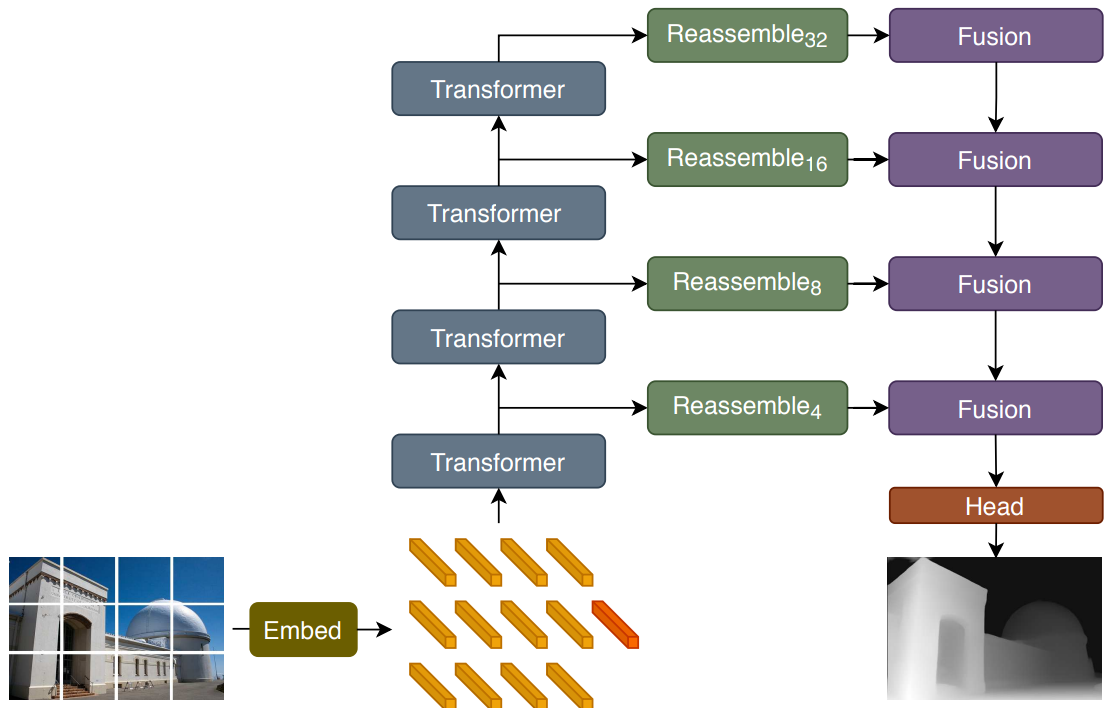}
        \caption{}
    \end{subfigure}
    \hspace{\fill}
    \begin{subfigure}{0.45\linewidth}
        \includegraphics[width=\linewidth]{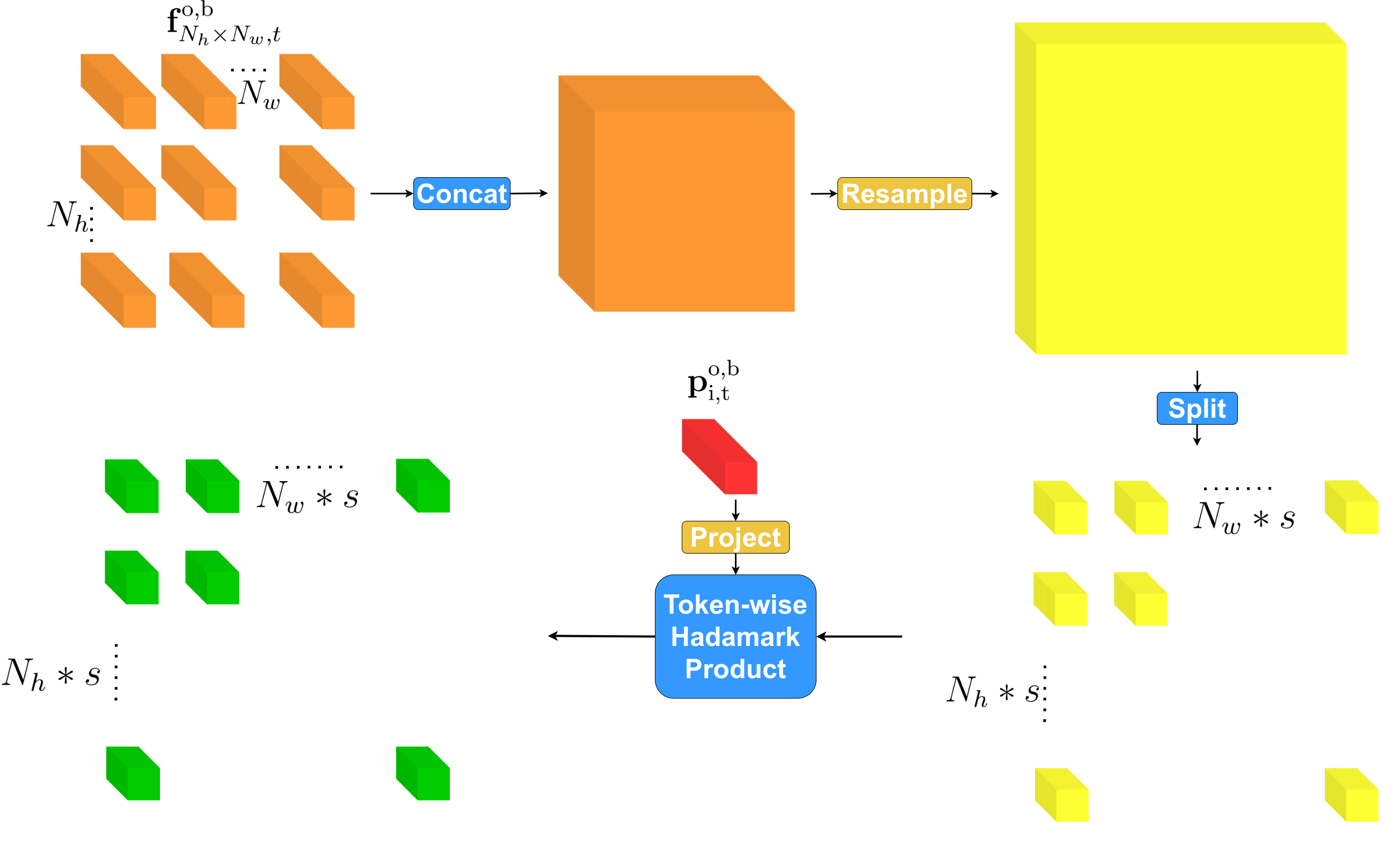}
        \caption{}
    \end{subfigure}
    \caption{The standard DPT (a, taken from~\cite{ranftl2021vision_dpt}) and our proposed person-conditioned re-assemble stage (b). This transformed DPT is used for predicting gaze heatmaps for each person in the scene.}
    \label{fig:conditioned-dpt}
    \vspace{-2mm}
\end{figure}

As mentioned in Section~\ref{sec:architecture}, we rely on the standard DPT~\cite{ranftl2021vision_dpt} decoder that has been developed for dense prediction, 
and propose an interesting  way to transform it  
for performing person-conditioned gaze heatmap prediction.

Similar to an FPN~\cite{lin2017fpn}, the DPT assembles the set
of ViT image tokens into image-like feature representations at various
resolutions. The feature representations are then progressively
fused into the final dense prediction. Specifically, the DPT decoder contains two stages:
1) A \textit{Re-assemble} stage to construct feature maps at specific resolutions at each block, and 2) a \textit{Fusion} stage where the feature maps across consecutive blocks are upscaled and combined. 

To include the person specific information, as represented by the person token in our architecture, we modify the re-assemble stage to filter only the information relevant for the given person (Figure~\ref{fig:conditioned-dpt}). 
More precisely, following the standard DPT, we first project the input frame tokens $\outimgtokens_{t}$ at level $b$ to a lower token dimension using a $1 \times 1$ conv layer, followed by spatial upsampling or downsampling using a strided $3 \times 3$ transposed conv layer or conv layer respectively;
\begin{equation}
    \imgtokens^{\text{DPT,b}}_{t} = \text{Split} (\text{Resample}^b ( \text{Concat} ( \outimgtokens_{t} )))
\end{equation}
where Split is the reverse of the spatial concatenation operation (Concat), and Resample is the spatial upsampling/downsampling operation. To condition on a person, we then perform a token-wise hadamard product of the projected frame tokens with the the projected person token from the corresponding block $b$.
\begin{equation}
    \imgtokens^{\text{DPT-c,b}}_{t} = \imgtokens^{\text{DPT,b}}_{t} * \dptproj^b ( \outpersontoken_{i,t} )
\end{equation}
where $*$ denotes the hadamard product. The standard Fusion stage then follows to obtain the predicted gaze heatmap.

%
%

\end{document}